\documentclass[]{new_tlp}
\usepackage[]{xcolor}
\usepackage{mathptmx}
\usepackage{amsmath}\usepackage{amsfonts}
\usepackage{hyperref}

\newcommand{\longversion}[1]{}

\usepackage{tikz}
\usetikzlibrary{fit,positioning,arrows,shapes,snakes,automata,%
                petri,%
                topaths,calc}

\colorlet{vertexTopColor}{white}
\colorlet{vertexBottomColor}{black!10}

\tikzstyle{tdnode} = [draw,rounded corners,top color=vertexTopColor,bottom color=vertexBottomColor,minimum size=1.5em]
\tikzstyle{stdnode} = [tdnode, font=\scriptsize]
\tikzstyle{stdnodecompact} = [stdnode, inner sep = 1.5pt, outer sep = 0.1pt]
\tikzstyle{stdnodetable} = [stdnode, inner sep = 1.5pt, outer sep = 0]
\tikzstyle{stdnodenum} = [minimum size=1.5em, font=\scriptsize]
\tikzstyle{tdedge} = [-,draw,thick]
\tikzstyle{tdlabel} = [draw=none, rectangle, fill=none, inner sep=0pt, font=\scriptsize]

\tikzstyle{squigarrow} = [->,line join=round,decorate, decoration={
    zigzag,segment length=4,amplitude=.9,post=lineto,post length=2pt}]

\newcommand{\SB}{\{}%
\newcommand{\SE}{\}}%
\newcommand{\eqdef}{\ensuremath{\,\mathrel{\mathop:}=}}

\newcommand{\FIX}[1]{#1} 

\newcommand{\tab}[1]{\ensuremath{\tau_{#1}}}
\newcommand\bcmdtab{\noindent\bgroup\tabcolsep=0pt%
  \begin{tabular}{@{}p{10pc}@{}p{20pc}@{}}}
\newcommand\ecmdtab{\end{tabular}\egroup}

\usepackage{url}

  \title[Utilizing Treewidth for Quantitative Reasoning on Epistemic Logic Programs]
        {Utilizing Treewidth for Quantitative Reasoning on Epistemic Logic Programs}

  \author[Besin, Hecher, and Woltran]
         {VIKTOR~BESIN\\
         TU Wien, Vienna, Austria\\
         \email{vbesin@dbai.tuwien.ac.at}\\
         MARKUS HECHER\\
         TU Wien, Vienna, Austria\\
         \email{hecher@dbai.tuwien.ac.at}\\
         STEFAN WOLTRAN\\
         TU Wien, Vienna, Austria\\
         \email{woltran@dbai.tuwien.ac.at}}

\jdate{March 2003}
\pubyear{2003}
\pagerange{\pageref{firstpage}--\pageref{lastpage}}
\doi{S1471068401001193}

\newtheorem{definition}{Definition} 
\newtheorem{example}{Example} 
\newtheorem{proposition}{Proposition} 

\newcommand{\algo}[1]{\ensuremath{\mathbb{#1}}}

\newcommand{\tuplecolor}[1]{\textcolor{#1}}
\newcommand{\inputPredColor}{orange!55!red}

\newcommand{\epistemiccolor}{green!62!black}
\newcommand{\specialPredColor}{red!62!black}

\usepackage{multirow}   
\usepackage{booktabs}   
\newcommand{\ATab}[1]{\ensuremath{\text{Comp}}}

\newcommand{\ax}{\textsf{a}}
\newcommand{\ex}{\textsf{e}}
\def\hy{\hbox{-}\nobreak\hskip0pt}
\newcommand{\var}{\ax{\hy}ats}
\newcommand{\uvar}{ats}
\newcommand{\vvar}{{\hy}ats}
\newcommand{\evar}{\ex{\hy}ats}

\newcommand{\prob}{\mathsf{prob}}

\DeclareMathOperator{\compat}{comp}
\newcommand{\primal}[1]{\ensuremath{G_{#1}}}
\newcommand{\epistemic}[1]{\ensuremath{E_{#1}}}
\newcommand{\nested}[2]{\ensuremath{G_{#1}^{#2}}}

\newcommand{\Card}[1]{|#1|}
\DeclareMathOperator{\depth}{depth}

\DeclareMathOperator{\post}{post-order}

\newcommand{\hdpa}{\ensuremath{\mathtt{NestELP}}\xspace}
\newcommand{\adpa}{\ensuremath{\mathtt{HybDP}}\xspace}
\newcommand{\dpa}{\ensuremath{\mathtt{DP}}\xspace}
\newcommand{\algEASP}{\algo{\#ELP}}%
\newcommand{\algQELP}{\algo{PELP}}%

\newcommand{\cELP}{\#\textsc{ELP}\xspace}

\newcommand{\nesthdb}{\texttt{nesthdb}\xspace}
\newcommand{\nestelp}{\texttt{nestelp}\xspace}
\newcommand{\eclingo}{\texttt{eclingo}\xspace}
\newcommand{\clingo}{\texttt{clingo}\xspace}
\newcommand{\htd}{\texttt{htd}\xspace}

\DeclareMathOperator{\type}{type}
\newcommand{\intr}{\textsf{intr}}
\newcommand{\leaf}{\textsf{leaf}}
\newcommand{\rem}{\textsf{rem}}
\newcommand{\join}{\textsf{join}}
\DeclareMathOperator{\width}{width}
\newcommand{\tw}[1]{\mathit{tw}(#1)}
\DeclareMathOperator{\rootOf}{root}
\DeclareMathOperator{\children}{children}

\newcommand{\eneg}{\mathbf{not}\xspace}
\newcommand{\kop}{\mathbf{K}\xspace}
\newcommand{\mop}{\mathbf{M}\xspace}

\newcommand{\answersets}[1]{AS(#1)}
\newcommand{\wvs}[1]{WVS(#1)}

\newcommand{\progASP}{\mathsf{P}}
\newcommand{\nop}{\neg}
\newcommand{\prog}{\Pi}
\newcommand{\Tab}[1]{\ensuremath{\text{Child-Tabs}}}

\makeatletter
\let\O@argtabularcr\@argtabularcr
\def\O@xtabularcr{\@ifnextchar[\O@argtabularcr{\ifnum 0=`{\fi}\cr}}
\let\O@tabacol\@tabacol
\let\O@tabclassiv\@tabclassiv
\let\O@tabclassz\@tabclassz
\let\O@tabarray\@tabarray
\def\author@tabular{\authorsize\def\@halignto{}\@authortable}
\let\endauthor@tabular=\endtabular
\def\author@tabcrone{{\ifnum0=`}\fi\O@xtabularcr\affilsize\itshape
 \let\\=\author@tabcrtwo\ignorespaces}
\def\author@tabcrtwo{{\ifnum0=`}\fi\O@xtabularcr[-3\p@]\affilsize\itshape
 \let\\=\author@tabcrtwo\ignorespaces}
\def\@authortable{\leavevmode \hbox \bgroup $\let\@acol\O@tabacol
 \let\@classz\O@tabclassz \let\@classiv\O@tabclassiv
 \let\\=\author@tabcrone \ignorespaces \O@tabarray}
\makeatother

\usepackage{array}
\newcolumntype{H}{>{\setbox0=\hbox\bgroup}c<{\egroup}@{}}
\usepackage[ruled,vlined,linesnumbered]{algorithm2e}

\SetKwInput{KwData}{In}
\SetKwInput{KwResult}{Out}
\SetAlFnt{\small}
\SetAlCapFnt{\small}
\SetAlCapNameFnt{\small}
\SetAlCapHSkip{0pt}
\SetEndCharOfAlgoLine{}
\IncMargin{-0.4em}
\makeatletter
\newcommand{\algorithmfootnote}[2][\footnotesize]{
  \let\old@algocf@finish\@algocf@finish
  \def\@algocf@finish{\old@algocf@finish
    \leavevmode\rlap{\begin{minipage}{\linewidth}
    #1#2
    \end{minipage}}
  }
}
\makeatother

\SetNlSkip{.25em}

\newcommand{\TTT}{\mathcal{T}}
\newcommand{\calI}{\mathcal{I}}
\newcommand{\calA}{\mathcal{A}}

\begin{document}


\maketitle
  \begin{abstract}
  Extending the popular  Answer Set Programming (ASP) paradigm by introspective reasoning capacities has received increasing interest within the last years. Particular attention is given to the formalism of epistemic logic programs (ELPs) where standard rules are equipped with modal operators which allow to express conditions on 
	  literals for being known or possible, i.e., contained in all or some answer sets, respectively. ELPs thus deliver multiple collections of answer sets, known as world views.
%
Employing ELPs for reasoning problems so far has mainly been restricted to 
standard decision problems (complexity analysis) and enumeration (development of systems) 
	  of world views.
  %
  %
  %
In this paper, we take a next step and contribute to epistemic logic programming in two ways:
	  First,
	  we establish quantitative reasoning for ELPs, where the acceptance of a certain set of literals 
	  depends on the number (proportion) of world views that are compatible with the set.
	  Second,
  we present a novel system that is capable of efficiently solving the underlying counting problems required to answer such quantitative reasoning problems.
  Our system exploits the graph-based measure treewidth and works by iteratively finding and refining (graph) abstractions of an ELP program. On top of these abstractions, we apply dynamic programming that is combined with utilizing existing search-based solvers like (e)clingo for hard combinatorial subproblems that appear during solving.
  It turns out that our approach is competitive with existing systems that were introduced recently. This work is under consideration for acceptance in TPLP.
  \end{abstract}



\section{Introduction}

\emph{Answer Set Programming (ASP)} 
is a well-studied problem modeling and solving
framework that is particularly suited for solving problems related to knowledge representation and reasoning and artificial intelligence, see, e.g.,~\cite{BrewkaEiterTruszczynski11}\longversion{\cite{GebserEtAl19,AlvianoDodaro20,AbelsEtAl19}}.
\FIX{In ASP}, questions are modeled in the form of \emph{logic programs (LPs)},
which can be seen as a rule-based language whose solutions are referred to by \emph{answer sets} and which has been significantly extended over the time\longversion{~\cite{CalimeriEtAl20}}.
The major driver in enabling the use of logic programs for a broad use in both academia and industry was the development of efficient solvers. 
However, while the ASP framework is quite powerful, its limits in terms of expressiveness are visible when turning the attention to epistemic specifications.
The idea of these epistemic specifications, which dates back to the early 90s~\longversion{\cite{lpnmr:GelfondP91}}\cite{aaai:Gelfond91}, allows to precisely describe the behavior of rational agents who are capable of reasoning  over multiple worlds. There, depending on whether some objections are possible (true in some world) or known (i.e., true in all worlds) certain consequences have to be derived.
This is often modeled by means of operators $\mathbf{K}$ or
$\mathbf{M}$, which represents that certain objections are \emph{known to be true} or are
\emph{possibly true}, respectively. 
Internally these operators can be translated to \emph{epistemic negation}~$\eneg$, which expresses that some objection is \emph{not known}, i.e., not true in all worlds.
%
Enhancing standard rules by such operators leads
to the development of \emph{epistemic logic programs (ELPs)}. 
%
Indeed, depending on the different semantics for ELPs, which have been developed and refined over the years, e.g.,~\longversion{\cite{lpnmr:Gelfond11,ijcai:CerroHS15}}\cite{
Truszczynski11,logcom:KahlWBGZ15,ai:ShenE16,lpnmr:CabalarFC19}, usual reasoning problems like \emph{world view existence} and certain extensions reach the third and fourth level of the polynomial hierarchy, respectively, and thus
are considered significantly harder than reasoning in standard ASP~\cite{EiterGottlob95}.

In this work, we take a step further and initiate the study of \emph{quantitative reasoning} for ELPs, where decisions concerning the acceptance of a given set of literals depend on the actual \emph{number (proportion) of world views} compatible with the set.
This allows us to reason about the acceptance of certain literals based on the likelyhood of being contained in an arbitrary world view.
\longversion{\begin{example}
	Consider an ELP problem encoding with a number of world views, e.g. program~$\prog$ has eight world views. Further consider special interest in specific atoms $a$ and $b$ with the desire of $a$ and $\neg b$, which are part of the result in the majority of the world views, e.g.~$a$ and $\neg b$ hold in seven world views of~$\prog$. With the resulting probability of~$\frac{7}{8}$ one could argue that it is reasonable to accept~$a$ and~$\neg b$ at the same time, since this outcome is very likely and the inverse is unlikely. 
\end{example}}
To the best of our knowledge, a few works on quantitiative reasoning in ASP exist, e.g.,~\cite{FierensEtAl15}, but it has not yet been studied for ELPs. 
As a second contribution we present a new system tailored for quantitiative reasoning in ELPs.
Although there has been progress in developing ELP solvers 
%
(e.g., EP-ASP~\cite{ijcai:SonLKL17}, selp~\cite{BichlerMorakWoltran20} and a very recent extension of clingo for epistemic logic programs, called eclingo~\cite{CabalarEtAl20}), these approaches basically rely on reducing ELP problems
to standard ASP. Thus, these solvers typically materialize all world views, which is not necessary for quantitative reasoning.
We take here a novel route
%
by utilizing ideas from parameterized algorithmics which appear better
suited for counting problems that underly the quantitative reasoning approach.

\FIX{Our approach works on abstractions of the internal (graph) structure of ELPs; i.e., we take the \emph{primal graph}\footnote{Basically, the primal graph of an (E)LP comprises of the atoms of the program, where two atoms are adjoined by an edge whenever these two atoms appear together in at least one rule.} of an ELP and contract certain paths between
nodes referring to epistemic literals.
On this graph we implicitly utilize the measure \emph{treewidth}, which
aims at measuring the tree-likeness of a given graph.}
The measure treewidth gives rise to a so-called \emph{tree decomposition},
which allows to solve a problem by following a divide-and-conquer approach, where world views of ELPs are computed by solving subprograms and combining world views accordingy.
Our solver adheres to this approach, where we approximate suitable abstractions of the primal graph structure of an ELP in order to evaluate the program in a way that is guided along a tree decomposition of the abstraction. \FIX{So, the idea of these abstractions compared to the full primal graph is to decrease treewidth such that still structural information in the form of tree decompositions can be utilized.}
In addition to the abstractions and in order to efficiently apply our approach also to (practical) ELPs of high treewidth, we present the following additions: (i) We nest the computation of abstractions and (ii) for hard combinatorial subprograms of (E)LPs, we employ existing standard solvers like (e)clingo. Both additions combined, together with the guidance of abstract (implicit) structure of ELPs, allows us to efficiently evaluate ELPs.

\vspace{-1em}
\paragraph{Contributions.} More concretely, we establish the following.
\begin{enumerate}
	\item We motivate the problem of \emph{world view counting}. This then leads to \emph{probabilistic world view acceptance}, which accepts certain literals based on a quantitative argument concerning the proportion of world views agreeing with those literals.
	\item Rooted in the theoretical investigation of~\cite{HecherMorakWoltran20}, we take up this idea and design an improved algorithm for evaluating epistemic logic programs by means of treewidth. 
	Our algorithm lifts nested dynamic programming~\cite{HecherThierWoltran20} from satisfiability to logic programming, where treewidth is utilized on subsequently refined abstractions.
	\item Finally, we present a system that implements this algorithm for 
		quantitative reasoning. It turns out that the system is competitive and scales 
		well for typical benchmarks.  
\end{enumerate}

\vspace{-1em}
\paragraph{Related Work.}
Treewidth was already utilized for the evaluation of standard LPs, e.g.,~\cite{ijcai:JaklPW09,Hecher20}\longversion{\cite{lpnmr:FichteHMW17}}.
The concept of using abstractions was stipulated before as well, but in a different context~\cite{HecherThierWoltran20} or with the purpose of establishing theoretical results~\cite{stacs:GanianRS17}\longversion{\cite{HecherMorakWoltran20,DellRothWellnitz19,EibenEtAl19}}.
However, we improved an existing algorithm~\cite{HecherMorakWoltran20} and to the best of our knowledge, our solver is the first implementation of solving ELPs that is guided by tree decompositions. While the solver 
selp~\cite{BichlerMorakWoltran20} 
uses decompositions for breaking large rules into smaller ones, the solving itself is not guided by tree decompositions.
Also studies for measures different from treewidth have been conducted in the ASP domain, e.g.,~\cite{tocl:LoncT03,ecai:BliemOW16,amai:FichteKW19}\longversion{\cite{lpnmr:FichteH19,ai:GottlobPW10,ai:FichteS15}}.
%

\section{Preliminaries}


\paragraph{Answer Set Programming (ASP).}

We 
follow standard definitions of propositional ASP~\cite{BrewkaEiterTruszczynski11}.
%
Let $k$, $m$, $n$ be non-negative integers such that
$k \leq m \leq n$ and $a_1$, $\ldots$, $a_n$ be distinct propositional
atoms. Moreover, we refer by \emph{literal} to an atom or the negation
thereof.
%
A \emph{program}~$\progASP$ is a set of \emph{rules} of the form
%
\(
a_1\vee \cdots \vee a_k \leftarrow a_{k+1}, \ldots, a_{m}, \neg
a_{m+1}, \ldots, \neg a_n.
\)

For a rule~$r$, we let $H_r \eqdef \{a_1, \ldots, a_\ell\}$,
$B^+_r \eqdef \{a_{\ell+1}, \ldots, a_{m}\}$, and
$B^-_r \eqdef \{a_{m+1}, \ldots, a_n\}$.
%
%
%
We denote the sets of \emph{atoms} occurring in a rule~$r$ or in a
program~$\progASP$ by $\uvar(r) \eqdef H_r \cup B^+_r \cup B^-_r$ and
$\uvar(\progASP)\eqdef \bigcup_{r\in\progASP} \uvar(r)$.
An \emph{interpretation} $I\subseteq \uvar(\progASP)$ is a set of atoms. $I$ \emph{satisfies} a
rule~$r$ if $(H_r\,\cup\, B^-_r) \,\cap\, I \neq \emptyset$, or
$B^+_r \setminus I \neq \emptyset$\FIX{, or both}.  $I$ is a \emph{model} of $\progASP$
if it satisfies all rules of~$\progASP$. 
%
%
The \emph{Gelfond-Lifschitz
  (GL) reduct} of~$\progASP$ under~$I$ is the program~$\progASP^I$ obtained
from $\progASP$ by first removing all rules~$r$ with
$B^-_r\cap I\neq \emptyset$ and then removing all~$\neg z$ where
$z \in B^-_r$ from the remaining
rules~$r$\longversion{~\cite{GelfondLifschitz91}}. %
Then, $I$ is an \emph{answer set} of a program~$\progASP$ if $I$ is a minimal
model of~$\progASP^I$. %
We refer to the set of answer sets of a given program~$\progASP$ by~$\answersets{\progASP}$.
%
The problem of deciding whether a program has an answer set, i.e., whether~$\answersets{\progASP}\neq \emptyset$, 
is $\Sigma_{2}^{P}$-complete~\cite{EiterGottlob95}. 
%
%

\begin{example}\label{ex:running} 
\vspace{-1em} 
	Consider the program~
	$\progASP \eqdef \SB 
	\overbrace{\{ a\lor b\}}^{r_1},\;
	\overbrace{\{c \leftarrow  \nop d\}}^{r_2},\;
	\overbrace{\{d \leftarrow \nop c\}}^{r_3} \SE$.
	The set $AS(\progASP)$, denoting the answer sets for the logic program $\progASP$, consists of $\{a,c\}$, $\{a,d\}$, $\{b,c\}$ and $\{b,d\}$.
\end{example}

\paragraph{Tree Decompositions and Treewidth.} %
We assume that graphs are undirected, simple, and free of self-loops.
%
Let~$G=(V,E)$ be a graph and~$U\subseteq V$ be a set of vertices.
Then, $G-U\eqdef (V\setminus U, \{e\in E\mid e\cap U=\emptyset\})$ is the graph obtained from removing~$U$ from $G$. 
Further, $U$ is a \emph{connected component} of a graph~$G'=(V',E')$ if
$U\subseteq V'$, $U$ is \emph{connected} and~$U=\{u'\mid u\in U, \{u,u'\}\in E'\}$.

 Let $G =
(V, E)$ be a graph, $T$ a rooted tree with \emph{root node~$\rootOf(T)$}, and $\chi$ a labeling function that maps
every node $t$ of $T$ to a subset $\chi(t) \subseteq V$ called the \emph{bag} of $t$. The pair
$\mathcal{T} = (T, \chi)$ is called a \emph{tree decomposition (TD)}~\cite{BodlaenderKloks96}\longversion{\cite{RobertsonSeymour86}} of $G$
iff (i) for each $v \in V$, there exists a $t$ in $T$, such that $v \in
\chi(t)$; (ii) for each $\{v, w\} \in E$, there exists $t$ in $T$, such that
$\{v, w\} \subseteq \chi(t)$; and (iii) for each $r, s, t$ of $T$, such that $s$
lies on the unique path from $r$ to $t$, we have $\chi(r) \cap \chi(t) \subseteq
\chi(s)$. 
\FIX{Intuitively, a tree decomposition allows to solve problems on a graph by analyzing parts of the graph and combining solutions to these accordingly. In order to simplify presentation, restricted node types and decompositions are oftentimes used, which are given as follows.}
For a node~$t$ of~$T$, we say that $\type(t)$ is $\leaf$ if $t$ has
no children and~$\chi(t)=\emptyset$; $\join$ if $t$ has children~$t'$ and $t''$ with
$t'\neq t''$ and $\chi(t) = \chi(t') = \chi(t'')$; $\intr$
(``introduce'') if $t$ has a single child~$t'$,
$\chi(t') \subseteq \chi(t)$ and $|\chi(t)| = |\chi(t')| + 1$; $\rem$
(``removal'') if $t$ has a single child~$t'$,
$\chi(t') \supseteq \chi(t)$ and $|\chi(t')| = |\chi(t)| + 1$. 
If for
every node $t\in T$, %
$\type(t) \in \{ \leaf, \join, \intr, \rem\}$, 
then $(T,\chi)$ is called \emph{nice}.
\FIX{For every TD, one can compute a nice tree decomposition in polynomial time~\cite{BodlaenderKloks96} without increasing the width by adding intermediate (auxiliary) nodes accordingly.}
%
%
The \emph{width} of a TD is defined as the cardinality of its largest bag minus
one. The \emph{treewidth} of a graph $G$, denoted by $\tw{G}$, is the minimum
width over all TDs of $G$.  Note that if $G$ is a tree, then $\tw{G} = 1$.


\longversion{
\paragraph{Tree Decompositions for Logic Programs.}
In order to use tree decompositions for logic programs, we require a graph representation. To this end, we define the so-called \emph{primal graph}~$\primal{\progASP}$ of a logic program~$\progASP$, whose vertices consist of the atoms~$\var(\progASP)$ and there is an edge between two vertices whenever the corresponding atoms appear together in at least one common rule of~$\progASP$. Formally, we let~$\primal{\progASP}=(\var(\progASP), E)$ with~$E$ being~$\{\{a,b\}\subseteq\var(r)\mid r\in\progASP\}$.

\begin{example}
	In Figure~\ref{fig:graph-td} the primal graph $G_{\progASP}$ for $\progASP$ as defined in Example~\ref{ex:running} as well as one corresponding TD~$\TTT$ of $G_{\progASP}$ of width 3 can be seen. Notice that for the primal graph $G_{\prog}$ for the epistemic logic program $\prog$ from Example~\ref{ex:running2} $G_{\prog} = G_{\progASP}$ holds.
\end{example}

Let~$\progASP$ be an logic program, $\mathcal{T}=(T,\chi)$ be a TD of~$\primal{\progASP}$, and~$t$ be a node of~$T$. 
Then, we let the \emph{bag program~$\progASP_t\eqdef\{r\in\progASP \mid \var(r)\subseteq\chi(t)\}$}.
}

\section{Counting and Reasoning for Epistemic Programs}


\paragraph{Epistemic Logic Programming.}

An \emph{epistemic literal} is a formula
$\eneg \ell$, where $\ell$ is a literal and $\eneg$ is the epistemic negation
operator. 
Let $k$, $m$, $j$, $n$ be non-negative integers such that
$k \leq m \leq j \leq n$ and $a_1$, $\ldots$, $a_n$ be distinct propositional
atoms.
An \emph{epistemic logic program (ELP)} is a set $\Pi$ of
\emph{ELP rules} of the form
%
  %
   $a_1\vee \cdots \vee a_k \leftarrow \ell_{k+1}, \ldots, \ell_m, \xi_{m+1}, \ldots,
   \xi_j, \neg \xi_{j + 1}, \ldots, \neg \xi_{n}$,
  %
%
where each $\ell_i$ with~$k+1\leq i\leq m$ is a literal over atom~$a_i$, and each $\xi_i$ with~$m+1\leq i\leq n$ is an {epistemic literal} of the form $\eneg 
\ell_i$, where $\ell_i$ is a literal over atom~$a_i$. 
%
%
%
Then, $\uvar(r)\eqdef\{a_1,\ldots,a_n\}$ denotes the set of atoms ocurring in an
ELP rule $r$, $\evar(r) \eqdef \{a_{m+1},\ldots,a_n\}$ denotes the
set of \emph{epistemic atoms}, i.e., those used in epistemic literals of $r$, 
and $\var(r)\eqdef \uvar(r)\setminus\evar(r)$ refers to the \emph{non-epistemic} atoms of~$r$. We call~$r$ \emph{purely-epistemic} if~$\var(r)=\emptyset$. These notions naturally
extend to programs.
In a rule we sometimes write $\kop \ell$ and $\mop \ell$ for a literal~$\ell$, which refers to the expressions~$\neg \eneg \ell$ and~$\eneg \neg \ell$, respectively.

\longversion{
In order to define the semantics of an ELP, we will use the approach by
\citeauthor{iclp:Morak19} \shortcite{iclp:Morak19}, which follows the semantics
defined in \cite{ai:ShenE16}, but uses a different formal
representation. Note that, however, our results can be adapted to other
``reduct-based'' semantics, by changing the definition of the reduct
appropriately.}
Given an ELP $\Pi$, a \emph{world view interpretation
(WVI)} $I$ for $\Pi$ is a consistent set $I$ of literals over a set~$A\subseteq \uvar(\prog)$ of atoms, i.e., $I\subseteq \{a,\neg a\mid a\in A\}$ such that there is no~$a\in A$ with~$\{a,\neg a\}\subseteq I$. 
Intuitively, every~$\ell\in I$ is considered as ``known'' \FIX{and every~$a\in A$ with~$\{a,\neg a\}\cap I=\emptyset$ is treated as ``possible''}. 
We denote the WVI over a set~$X\subseteq \uvar(\prog)$ of atoms obtained by restricting~$I$ to~$Y=(A\cap X)$ by~$I_{|X}\eqdef I\cap \{a,\neg a\mid a\in Y\}$.
%
Next, we define compatibility with a set of interpretations. 

\begin{definition}[WVI Compatibility]\label{def:compat}
  Let $\calI$ be a set of interpretations over a set $A$ of atoms. Then, a
  WVI $I$ is \emph{compatible} with $\calI$ if:
  \vspace{-.5em}
  \begin{enumerate}
    \item\label{def:compat:1} $\calI \neq \emptyset$;
    \item\label{def:compat:2} for each atom $a \in I$, it holds that for each
		  $J \in \calI$, $a \in J$;
    \item\label{def:compat:3} for each $\neg a \in I$, we have for each
	    	$J \in \calI$, $a \not\in J$;
    \item\label{def:compat:4} for each atom $a \in A$ with~$\{a,\neg a\}\cap I=\emptyset$, there are
	    $J, J' \in \calI$, such that $a \in J$, but $a \not\in J'$.
  \end{enumerate}
\end{definition}

While there are many different semantics, e.g.,~\longversion{\cite{lpnmr:Gelfond11,ijcai:CerroHS15}}\cite{
aaai:Gelfond91,Truszczynski11,logcom:KahlWBGZ15,ai:ShenE16}, we follow the approach of~\cite{aaai:Gelfond91},
syntactically denoted according to recent work~\cite{iclp:Morak19}.
The \emph{epistemic reduct}~\cite{aaai:Gelfond91} of program $\Pi$ w.r.t.\ a WVI $I$ over~$A$, denoted $\Pi^I$, is defined as 
$\Pi^I = \{ r^I \mid r \in \Pi \}$ where $r^I$
denotes rule $r$ where each epistemic literal $\eneg \ell$, whose atom is also in~$A$, is replaced by
$\bot$ if $\ell \in I$, and by $\top$ otherwise.  Note that $\Pi^I$
is a plain logic program with all occurrences of epistemic negation
removed.  
%
Now, a WVI $I$ over~$\uvar(\prog)$ is a \emph{world view (WV)} of $\Pi$ iff 
$I$ is compatible with the set $\answersets{\Pi^I}$.
Without loss of generality we only consider ELPs~$\prog$, where every epistemic atom appears non-epistemically, i.e., $\evar(\prog)=\var(\prog^{\evar(\prog)})$.
We refer by~$\Pi\sqcup I$ to the ELP~$\Pi\cup\{ \leftarrow \neg \kop \ell \mid \ell\in I \} \cup \{\leftarrow \neg \mop a; \leftarrow \neg \mop \neg a \mid a\in A, a\notin I, \neg a\notin I\}$
used for verifying whether~$I$ can be extended to a WV.
The set of WVs of an
ELP $\Pi$ is denoted $\wvs{\Pi}$.
One of the reasoning tasks for ELPs is \emph{world view existence} deciding for an ELP $\Pi$ whether $\wvs{\Pi} \neq \emptyset$. This problem is known
to be $\Sigma_{3}^{P}$-complete \cite{Truszczynski11}. 
\longversion{Another interesting
reasoning task is deciding, given an ELP $\Pi = (\calA, \calR)$ and an
arbitrary propositional formula $\phi$ over $\calA$, whether $\phi$ holds in
some WV, that is, whether there exists $W \in \wvs{\Pi}$ such that $W \models
\phi$. This \emph{formula evaluation problem} is even harder, namely
\SIGMA{4}{P}-complete \cite{ai:ShenE16}.}

\begin{example}\label{ex:running2}
	Consider program $\prog \eqdef\progASP\cup
	\SB a \leftarrow\nop\kop b;\;\;
		b \leftarrow\nop\kop a;\;\;
		c \leftarrow\nop\kop d;\;\;
		d \leftarrow\nop\kop c;\;\;
		\leftarrow\nop\kop a,\nop\kop\nop a;\;\;
		\leftarrow\nop\kop b,\nop\kop\nop b;\;\;
		\leftarrow\nop\kop a,\nop\kop c;\;\;
		\leftarrow\nop\kop a,\nop\kop b,\kop c;\;\;
		\leftarrow\kop c,\kop d\SE$,
	where $\progASP$ is defined as in Example~\ref{ex:running}, i.e., the ELP $\prog$ depicts an epistemic extension of the plain logic program $\progASP$. For simplicity, let the rules be numbered equally from $r_1$ to $r_{12}$. When constructing a WVI~$I$ over~$\evar(\prog)$ one guesses for each atom~$a\in\evar(\prog)$ either (1)~$a\in I$, (2)~$\neg a\in I$ or (3)~$\{a,\neg a\}\cap I=\emptyset$ as described earlier, i.e., for the three atoms in~$\evar(\prog)$ we obtain~$3^4$ possibilities. Each WVI~$I$ can be checked with the corresponding epistemic reduct~$\Pi^I$ by verifying Definition~\ref{def:compat} for~$\answersets{\Pi^I}$.
	
	Consider~$I_1=\SB a, d, \neg b, \neg c\SE$ with its epistemic reduct~$\Pi^{I_1}\eqdef\progASP\cup
	\SB a;\;\;
	d\SE$.
	Note that the epistemic reduct is indeed a plain logic program, since by semantics of logic programs, rules $r$ with $\bot\in B^+_r$ or $\top\in B^-_r$ can obviously be dropped. Since~$\answersets{\Pi^{I_1}}=\SB\{a,d\}\SE$, compatibility of $I_1$ can be checked trivially which validates~$I_1$ as WV of~$\prog$. Similarly WVIs $I_2=\SB a, c, \neg b, \neg d\SE$ and $I_3=\SB b, c, \neg a, \neg d\SE$ can be constructed and correctly validated as WVs, i.e., $\wvs{\Pi}=\SB I_1, I_2, I_3\SE$.
\end{example}

\paragraph{Counting and Reasoning.}

In this work, we mainly cover the following counting
problem, which can then be used as a basis to solve (quantitative) reasoning problems.

\begin{definition}[World View Counting]\label{def:counting}
Let~$\prog$ be an ELP and~$Q$ be a WVI, called \emph{query}, over atoms~$\uvar(\prog)$. Then,
the problem~$\cELP(\prog, Q)$ asks to count the number of  world views~$W$ with~$Q\cap\uvar(\prog)\subseteq W$ and~$\{a\mid \neg a\in Q\}\cap W=\emptyset$. 
\end{definition}

As a special case, where~$Q=\emptyset$, a problem instance $\cELP(\prog, \emptyset)$ amounts to counting world views.
Interestingly, the problem can be used to reason about the likelihood of an atom or a set of atoms being contained in an arbitrary world view, defined as follows. 

\begin{definition}[Probability of World View Acceptance]\label{def:prob}
Let~$\prog$ be an ELP and~$Q$ be a WVI over $\uvar(\prog)$.
We define the \emph{probability of~$Q$} being compatible with a world view by~$\prob(\prog, Q)\eqdef\frac{\cELP(\prog, Q)}{\cELP(\prog, \emptyset)}$.
\end{definition}

Consequently, counting allows us to reason about the degree of believing in literals being part of world views.
This degree of belief can then be used for  accepting literals
depending on its probability exceeding a certain value, referred to by \emph{probabilistic world view acceptance}. 

\begin{example}
	Recall $\prog$ from Example~\ref{ex:running2}. Given~$Q\eqdef\{a,\neg b\}$, the number~$\cELP(\prog, Q) = 2$ naturally agrees with the number of WVs including~$a$, but not~$b$. The probability~$\prob(\prog, Q)= \frac{2}{3}$ can be used to argue about the chance of a WV of~$\prog$ containing~$a$ but not $b$, which renders~$a$ and~$\neg b$ very likely.
\end{example}

\noindent For Definitions~\ref{def:counting} and~\ref{def:prob}, we only consider
WVIs over epistemic atoms to simplify presentation.%
\footnote{This is not a hard restriction that could be circumvented for a non-epistemic atom~$a$, e.g., via constraint~$\leftarrow \neg\kop a, \kop a$.} 

\section{Quantitative Reasoning for ELPs via Dynamic Programming}

Next, we discuss core ideas of dynamic programming for the evaluation of epistemic logic programs.
We demonstrate this technique in Section~\ref{sec:dp} on a problem for ELPs that is much simpler than computing world views.
Then, we extend this technique to nested dynamic programming in order to count world views in Section~\ref{sec:ndp}, which finally leads to probabilistic reasoning. 

\subsection{Basics of Dynamic Programming}\label{sec:dp}
Algorithms that utilize treewidth for solving a problem in linear 
time typically proceed by \emph{dynamic
programming (DP)} along the tree decomposition.
Thereby, the tree is traversed in post-order and at each
node~$t$ of the tree, information is gathered~\cite{BodlaenderKloks96}
in a table~$\tau_t$.
A \emph{table}~$\tau_t$ is a set of rows, where a
\emph{row}~$u \in \tau_t$ is a sequence or tuple of fixed length. 
These tables are derived by an algorithm, which we therefore call
\emph{table algorithm}~$\algo{A}$.  
The actual length, content, and meaning of the rows depend on the
algorithm~$\algo{A}$ that derives tables.

The DP approach for solving problems of an epistemic logic program relies on a table algorithm~$\algo{A}$ and 
consists
of the following four steps:
\begin{enumerate}[leftmargin=*]
\item[\textbf{Prepare:}] Construct a \emph{graph representation}~$G$ of the given ELP~$\prog$. 
%
\item[\textbf{Decompose:}] Compute a tree decomposition~$(T,\chi)$ of~$G$, which can be obtained by using efficient heuristics~\cite{AbseherMusliuWoltran17a
}.
\item[\textbf{Compute:}]\label{step:dp} Execute table algorithm~$\algo{A}$ for every node~$t$ of~$T$ in post-order, 
  which returns the corresponding table for~$t$. 
\longversion{Algorithm~$\dpa$ works for given table algorithm~$\algo{A}$ as presented in Listing~\ref{fig:dpontd}
and takes as input instance~$\prog$ and a tree decomposition~$\TTT$.}
Algorithm~$\algo{A}$ takes as
  input 
    %
the corresponding bag~$\chi(t)$, the assigned instance~$\prog_t$ for node~$t$, as well as the child tables previously
  computed during the post-order traversal for child nodes of~$t$ in~$T$, and outputs a
  table~$\tau_t$. 
%
%
%
\item[\textbf{Output:}] Print the solution by interpreting the table for root~$n=\rootOf(T)$ of~$T$.
\end{enumerate}

\longversion{
\begin{algorithm}[t]%
  \KwData{%
    Table algorithm $\algo{A}$, program~$\progASP$, and a TD~$\TTT=(T,\chi)$ of the primal graph~$\primal{\progASP}$ with $n=\rootOf(T)$. 
  }%
  \KwResult{%
    Return whether~$\progASP$ admits an answer set by interpreting the table for node~$n$. 
  } %

  $\ATab{\algo{N}} \leftarrow \{\}\qquad \tcc*[h]{empty mapping}$

  \For{\text{\normalfont iterate} $t$ in $\post(T)$}{
    %
    %
    
    $\Tab{} \leftarrow \langle\ATab{\algo{N}}[t_1],\ldots, \ATab{\algo{N}}[t_\ell]\rangle\text{ where }\children(t)=\langle t_1,\ldots,t_\ell \rangle\hspace{-5em}$
  %
  
  $\ATab{\algo{A}}[t]  \leftarrow {\algo{A}}(\chi(t), \progASP_t, \Tab{})$
%
  }%
  \Return{$\ATab{\algo{A}}[n]\neq\emptyset\qquad \tcc*[h]{$\progASP$ admits an answer set iff $\ATab{\algo{A}}[n]\neq\emptyset$}$} 
 \caption{Algorithm ${\dpa}_{\algo{A}}(\progASP, \TTT)$ for computing models of~$\progASP$ via dynamic programming on
   TD~${\TTT}$.}
\label{fig:dpontd}
\end{algorithm}%
}

For simplicity and the ease of presentation, the table algorithms presented in this work are \emph{specified for nice tree decompositions} due to clear case distinctions depending on~$\type(t)$. \FIX{However, the implemented architecture \emph{does not} depend on certain normal forms of tree decompositions. So, our approach works independently of whether such a TD is nice or not, since the different cases can be combined programmatically and TD nodes of any interleaved (combined) type can be processed.}

Next, we briefly present a table algorithm for computing \emph{plausible world view interpretations} of an ELP~$\prog$,
which is a WVI~$I$ over~$\uvar(\prog)$ such that~$\answersets{\{r\in\prog\mid \var(r)=\emptyset\}^I}\neq\emptyset$, denoted by~$I\models_{\mathsf{p}} \prog$. 
Observe that every WV of~$\prog$ is always plausible as well.
While counting plausible WVI serves the purpose of demonstrating and explaining dynamic programming, interestingly it is actually a \#P-complete problem.

\begin{proposition}[Complexity of Counting Plausible WVIs] 
The problem of counting for a given ELP~$\prog$ the number of plausible WVIs is \#P-complete.
\end{proposition}\vspace{-.7em}
\begin{proof}[Proof (Sketch)]
For membership, observe that one can guess a WVI~$I$ and then check whether~$I\models_{\mathsf{p}}\prog$ in polynomial time.
Hardness is by reducing from \#SAT, where one aims for counting the number of models of a 3-CNF formula~$F=\{c_1,\ldots,c_l\}$. We construct an ELP~$\prog$ that contains for every variable~$v$ of~$F$ a rule~$\leftarrow \neg\kop v, \neg\kop \neg v$ and for every clause~$c_i=\ell_1 \vee \ell_2 \vee \ell_3$ of~$F$ a rule~$\leftarrow \neg \kop \ell_1, \neg \kop \ell_2, \neg \kop \ell_3$. Then, the number of plausible WVIs of~$\prog$ precisely captures the number of models of~$F$.
\end{proof}

\begin{figure}[t]%
	\centering
	\begin{tikzpicture}[node distance=7mm,every node/.style={fill,circle,inner sep=2pt}]
\node (ae) [label={[text height=1.5ex,yshift=0.0cm,xshift=0.05cm]left:$a^\ex$}] {};
\node (be) [right of=ae,label={[text height=1.5ex,yshift=0.0cm,xshift=-0.07cm]right:$b^\ex$}] {};
\node (ce) [below of=ae, label={[text height=1.5ex,yshift=0.0cm,xshift=0.05cm]left:$c^\ex$}] {};
\node (de) [right of=ce,label={[text height=1.5ex,yshift=0.0cm,xshift=-0.07cm]right:$d^\ex$}] {};

\draw (ae) to (ce);
\draw (ae) to (be);
\draw (be) to (ce);
\draw (ce) to (de);

\end{tikzpicture}%
	\qquad\qquad\qquad
	\includegraphics[scale=.85]{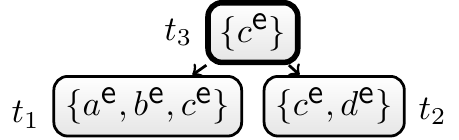}
	\vspace{-.75em}
	\caption{Epistemic primal graph~$\epistemic{\prog}$ (left) of~$\prog$ from Example~\ref{ex:running2} and a TD~$\mathcal{T}$ (right) of~$\epistemic{\prog}$.}
	%
	\label{fig:graph-td}%
\end{figure}
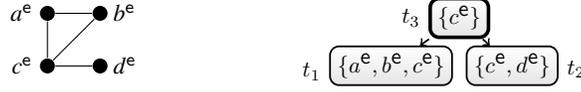

Before we discuss a table algorithm for counting plausible WVIs, we first require a \emph{graph representation}. To this end, we employ the \emph{epistemic primal graph}~$\epistemic{\prog}$ of an ELP~$\prog$, whose vertices stem only from the epistemic atoms $\evar(\prog)$ and there is an edge between two vertices whenever the corresponding epistemic atoms appear together in a common purely-epistemic rule of~$\prog$. Formally\footnote{For a set~$X$ of elements, we use the shortcuts~$X^\ex\eqdef\{x^\ex \mid x\in X\}$.}, 
we let~$\epistemic{\prog}=(\evar(\prog)^\ex, E)$ with~$E$ being~$\{\{a^\ex,b^\ex\} \mid  r\in\prog, \var(r)=\emptyset, \{a,b\}\subseteq\evar(r)\}$.
Now, let~$\mathcal{T}=(T,\chi)$ be a TD of the epistemic primal graph~$\epistemic{\prog}$ and~$t$ be a node of~$T$. Then, the \emph{epistemic bag program} for~$t$ is given by~$\prog_t\eqdef \{r \in \prog \mid \var(r)=\emptyset,\evar(r)^\ex \subseteq \chi(t)\}$. \FIX{This allows us to refer to the \emph{epistemic bag program up to~$t$} by~$\prog_{\leq t}\eqdef  \bigcup_{t'\text{ is a descendant node of }t\text{ in }T} \prog_{t'} \cup \prog_t$, which is the union over all epistemic bag programs for nodes below~$t$ in~$T$ Consequently, the epistemic bag program~$\prog_{\leq \rootOf(T)}$ up to the root corresponds to~$\prog$.}

\begin{example}\label{ex:running5}
Figure~\ref{fig:graph-td} depicts the epistemic primal graph~$\epistemic{\prog}$ for~$\prog$ as defined in Example~\ref{ex:running2} as well as one corresponding TD~$\TTT$ of~$\epistemic{\prog}$ of width 2. Further, consider the epistemic bag programs $\prog_{t_1} = \{r_8,r_9,r_{10},r_{11}\}$, $\prog_{t_2} = \{r_{12}\}$ and $\prog_{t_3} = \emptyset$. Note that by definition of~$\prog_t$ only rules solely built from $\evar(\prog)$, i.e., only purely-epistemic rules are being considered.
\FIX{Observe that for the root node~$t_3$ we have $\prog_{\leq t_3}=\prog$.}
\end{example}

Listing~\ref{fig:prim} depicts a table algorithm~$\#\algo{PWV}$ for counting plausible world view interpretations.  
Observe that it 
thereby suffices to compute WVIs over \emph{epistemic atoms}, as such a WVI already uniquely identifies one WVI over all atoms.
Then, algorithm~$\#\algo{PWV}$ stores rows of the form~$\langle I, c\rangle$, where~$I$ is a WVI over~$\chi(t)$ and \FIX{$c$ is an integer (counter) referring to the number of plausible WVIs of the epistemic bag program up to~$t$, that when restricted to~$\chi(t)$ coincide with~$I$. Consequently, for decompositions whose roots have empty bags, the counter of a stored row refers to the number of plausible world views of~$\Pi$.}
\FIX{As already mentioned above, for the ease of presentation, table algorithm~$\#\algo{PWV}$ is given for nice tree decompositions, i.e., in Listing~\ref{fig:prim} we distinguish the four different cases of nice TDs.}
So, if node~$t$ is a leaf node, cf.\ Line~\ref{lab:leaf}, the only row matching these conditions is~$\langle \emptyset, 0\rangle$.
Then, whenever a vertex~$a^\ex$ is introduced in a node~$t$, Line~\ref{lab:intr} guesses all three possibilities for extending an existing WVI~$I$ by atom~$a$ and checks that the resulting WVI~$J$ ensures~$\prog_t$.
For nodes~$t$ with~$\type(t)=\rem$, where we remove~$a^\ex$, Line~\ref{lab:rem} removes the mapping of~$a$ in any existing WVI~$I$ and sums up the counters of collapsing WVIs, i.e., where all atoms guessed in~$I'$ match, accordingly.
Finally for a join node~$t$, we intuitively keep only rows, whose WVIs are in all child nodes tables, and counters of those rows need to be multiplied. \FIX{Note that the clear case distinction between node types of nice TDs simplifies the processing of child tables, e.g., when processing a node of type $\join$, since there are at most two child nodes.}
%
%
 \begin{algorithm}[t]
   \KwData{Node~$t$, bag $\chi_t$, epistemic bag program~$\prog_t$, and child tables~$\langle \tau_1,\ldots,\tau_\ell\rangle$ of~$t$.}
   \KwResult{Table~$\tau_{t}.$}
   \uIf(\hspace{-1em})
   {$\type(t) = \leaf$}{%
     $\tau_t \leftarrow \{ \langle
     \tuplecolor{\inputPredColor}{\emptyset},  \tuplecolor{\epistemiccolor}{1}
     \rangle \}$ \label{lab:leaf}%
     %
   }%
  \uElseIf{$\type(t) = \intr$ and $a^\ex\in\chi_t$ is introduced}{
   \vspace{-0.05em}
   \makebox[3.34cm][l]{$\tau_{t} \leftarrow \{ \langle \tuplecolor{\inputPredColor}{J}, \tuplecolor{\epistemiccolor}{c} \rangle$}
     $|\;\langle \tuplecolor{\inputPredColor}{I}, \tuplecolor{\epistemiccolor}{c} \rangle \in \tau_1,  {{\tuplecolor{black}{J \in \{I, I \cup \{a\}, I \cup \{\neg a\}\}}, \tuplecolor{black}{J}}} \models_{\mathsf{p}} \prog_t
      \} \hspace{-5em}$\label{lab:intr}
     %
   \vspace{-0.05em}
     }\vspace{-0.05em}%
     \uElseIf{$\type(t) = \rem$ and $a^\ex \not\in \chi_t$ is removed}{%
       \makebox[3.45cm][l]{$\tau_{t} \leftarrow \{ \langle \tuplecolor{\inputPredColor}{I'}, \tuplecolor{\epistemiccolor}{\sum_{\langle J, c' \rangle\in\tau_1: I'\subseteq J} c'}
       \rangle$}$|\;\langle \tuplecolor{\inputPredColor}{I}, \tuplecolor{\epistemiccolor}{c}
       \rangle \in \tau_1, I'=I\setminus\{a,\neg a\} \}\hspace{-5em}$\label{lab:rem}
       \vspace{-0.1em}
     } %
     \uElseIf{$\type(t) = \join$}{%
       \makebox[2.2cm][l]{$\tau_{t} \leftarrow \{ \langle \tuplecolor{\inputPredColor}{I}, \tuplecolor{\epistemiccolor}{c_1\cdot c_2}
         \rangle$}$|\;\langle \tuplecolor{\inputPredColor}{I}, \tuplecolor{\epistemiccolor}{c_1} \rangle \in \tau_1, \langle\tuplecolor{\inputPredColor}{I}, \tuplecolor{\epistemiccolor}{c_2}\rangle\in \tau_2
       \}\hspace{-5em}$
       \vspace{-0.1em}
     } 
  \Return $\tab{t}$ \vspace{-0.25em}
     \caption{Table algorithm~$\#\algo{PWV}(\chi_t,\prog_t,\langle \tau_1,\ldots,\tau_\ell\rangle)$ for Counting Plausible WVIs.}
 \label{fig:prim}\label{alg:prim}
\end{algorithm}

\begin{figure}[t]%
\centering %
  \includegraphics{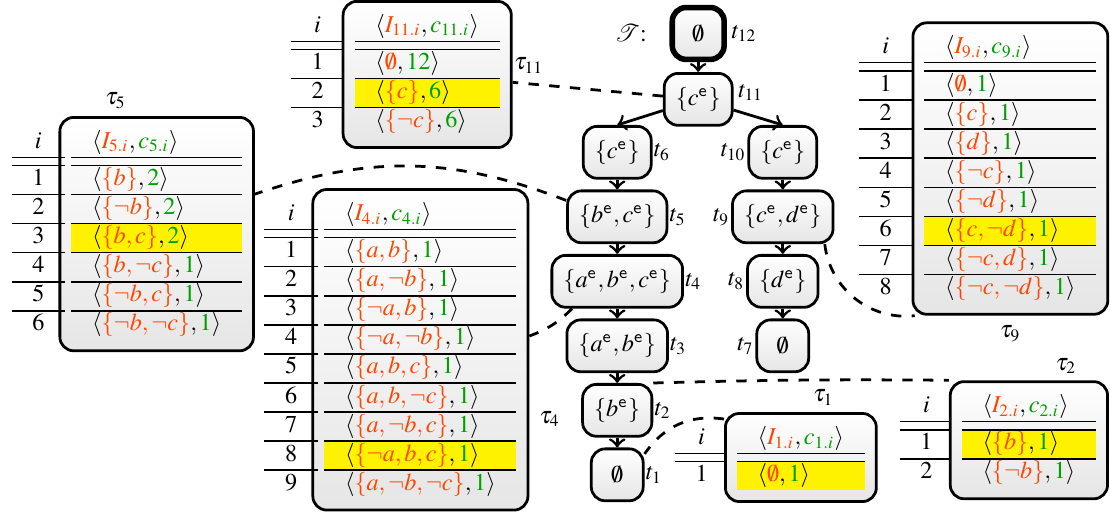}
\vspace{-0.75em}
\caption{
A nice TD~$\TTT$ of the epistemic primal graph~$\epistemic{\prog}$ of
  program~$\prog$ from Example~\ref{ex:running2} as well as
  selected 
  tables obtained by~$\#\algo{PWV}$ on~$\Pi$ and~$\TTT$.}
\label{fig:running1_prim}
\end{figure}%
\begin{example}\label{ex:sat}%
	Considering program~$\prog$ from Example~\ref{ex:running2}, we obtain three world views as described earlier. Table algorithm~$\#\algo{PWV}$ can be used to restrict the possible WVIs. Figure~\ref{fig:running1_prim} shows a nice tree decomposition~$\TTT=(T, \chi)$
  of~$\epistemic{\prog}$ and a selection of the tables~$\tab{1}$, $\ldots$, $\tab{12}$, which
  illustrate computation results obtained during post-order traversal
  of $\TTT$ by~$\#\algo{PWV}$.

  Table~$\tab{1}=\SB \langle\emptyset, 1\rangle\SE$ as per definition for
  $\type(t_1) = \leaf$.
  Since $\type(t_2) = \intr$, we construct table~$\tab{2}$
  from~$\tab{1}$ by taking~$I_{1.i}$, $I_{1.i}\cup \{b\}$ and $I_{1.i}\cup \{\neg b\}$ for
  each~$\langle I_{1.i}, c_{1.i} \rangle \in \tab{1}$ (corresponding to a guess on~$b$).
  Since $\evar(r_9)\subseteq \chi(t_2)$ we have $\prog_{t_2} = \{r_9\}$ for $t_2$ as described in Example~\ref{ex:running5}.
  In consequence, for each~$I_{2.i}$ of table~$\tab{2}$, we have
  $I_{2.i} \models \{r_9\}$ since~\algo{PWV} enforces
  satisfiability of $\prog_t$ in node~$t$.
  Then, $t_3$ introduces $a^\ex$ and $t_4$ introduces $c^\ex$ in similar fashion while satisfying the appropriate epistemic bag programs $\prog_{t_3} = \{r_8\}$ and $\prog_{t_4} = \{r_{10},r_{11}\}$.
  We derive tables~$\tab{7}$
  to $\tab{9}$ similarly.  Since $\type(t_5) = \rem$, we remove
  atom~$a$ from all elements in $\tab{4}$ to construct $\tab{5}$. As described earlier, this is accomplished by summing up the counters for matching WVIs when removing the atom~$a$, e.g., since the remaining, guessed atoms $b^e$ and $c^e$ are matching, counters for line 2 and 4 in table~$\tab{4}$ are summed up, resulting in line 2 in table~$\tab{5}$. Note
  that we have already seen all rules where $a^\ex$ occurs and hence $a^\ex$
  can no longer affect witnesses during the remaining traversal. We similarly
  construct
  $\tab{6}=\{\langle \emptyset,4 \rangle, \langle \{c\},3 \rangle, \langle\{\neg c\},2 \rangle\}$ and $\tab{{10}}=\{\langle \emptyset,3 \rangle, \langle \{c\},2 \rangle, \langle\{\neg c\},3 \rangle\}$.
  Since $\type(t_{11})=\join$, we construct table~$\tab{11}$ by taking
  the intersection $\tab{6} \cap \tab{{10}}$. Intuitively, this
  combines witnesses agreeing on~$c$ while multiplying the counters for matching guesses.
  Node~$t_{12}$ is again of type~$\rem$.
  By definition (primal graph and tree decompositions) for every~$r \in \prog$,
  atoms~$\var(r)$ occur together in at least one common bag.
  Hence, $\prog=\prog_{t_{12}}$ and since
  $\tab{12} = \{\langle \emptyset,24 \rangle \}$, we end up with 24 plausible WVIs of~$\prog$ which we can construct from the tables. For example, we obtain the
  interpretation~$\{\neg a,b,c,\neg d\} = I_{11.2} \cup I_{4.8} \cup I_{9.6}$,
as highlighted in yellow.
\end{example}%
%

\subsection{Counting World Views via Nested Dynamic Programming}\label{sec:ndp}

In order to extend DP for solving \cELP, 
we require a suitable graph representation that still allows for simple
table algorithms.
Let therefore~$\prog$ be an epistemic logic program.
Then, the \emph{primal graph~$\primal{\prog}$} uses atoms and epistemic atoms as vertices and it is defined by~$\primal{\prog}\eqdef (\{a^\circ \mid a\in\circ\vvar(\prog), \circ\in\{\ax,\ex\}\}, E)$, where~$E\eqdef \{\{a^\circ, b^\star\}\mid r\in\prog, a\in \circ\vvar(r), b\in\star\vvar(r),\{\circ,\star\}\subseteq\{\ax,\ex\}\} \cup \{\{a^\ax, a^\ex\} \mid a\in\evar(\prog)\}$.
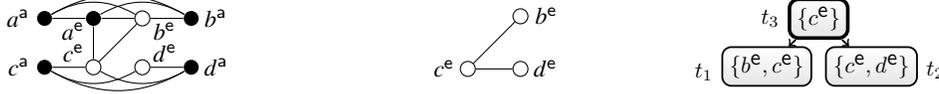
\begin{figure}[t]%
	\centering
%
\vspace{-.35em}
	\minipage{0.32\textwidth}
	\begin{tikzpicture}[node distance=6.5mm,every node/.style={fill,circle,inner sep=2pt}]
\node (a) [label={[text height=1.5ex,yshift=0.0cm,xshift=0.05cm]left:$a^\ax$}] {};
\node (ae) [right of=a,  label={[text height=1.5ex,yshift=-0.18cm,xshift=0.12cm]left:$a^\ex$}] {};
\node (be) [right of=ae,fill=white, draw=black,label={[text height=1.5ex,yshift=-0.18cm,xshift=-0.1cm]right:$b^\ex$}] {};
\node (b) [right of=be,label={[text height=1.5ex,yshift=0.0cm,xshift=-0.07cm]right:$b^\ax$}] {};

\node (c) [below of=a, label={[text height=1.5ex,yshift=0.0cm,xshift=0.05cm]left:$c^\ax$}] {};
\node (ce) [right of=c,fill=white, draw=black, label={[text height=1.5ex,yshift=0.18cm,xshift=0.12cm]left:$c^\ex$}] {};
\node (de) [right of=ce,fill=white, draw=black,label={[text height=1.5ex,yshift=0.18cm,xshift=-0.1cm]right:$d^\ex$}] {};
\node (d) [right of=de,label={[text height=1.5ex,yshift=0.0cm,xshift=-0.07cm]right:$d^\ax$}] {};

\draw (a) to (ae);
\draw (b) to (be);
\draw (c) to (ce);
\draw (d) to (de);

\draw[bend left] (a) to (b);
\draw[bend left] (a) to (be);
\draw[bend left] (ae) to (b);

\draw[bend right] (c) to (d);
\draw[bend right] (c) to (de);
\draw[bend right] (ce) to (d);

\draw (ae) to (ce);
\draw (ae) to (be);
\draw (be) to (ce);

\end{tikzpicture}%
	\endminipage\hfill
	\minipage{0.16\textwidth}
	\begin{tikzpicture}[node distance=7mm,every node/.style={fill,circle,inner sep=2pt}]
\node (b) [yshift=-2.5em,fill=white, draw=black, label={[text height=1.5ex,yshift=0.0cm,xshift=-0.05cm]right:$b^\ex$}] {};
\node (c) [below of=b,fill=white, draw=black, xshift=-7mm, label={[text height=1.5ex,yshift=0.0cm,xshift=0.07cm]left:$c^\ex$}] {};
\coordinate (middle) at ($ (b.south east)!.5!(c.south west) $);
\node (d) [right of=c,fill=white, draw=black, label={[text height=1.5ex,yshift=0.0cm,xshift=-0.07cm]right:$d^\ex$}] {};

\draw (b) to (c);
\draw (c) to (d);

\end{tikzpicture}
	\endminipage\hfill
	\minipage{0.32\textwidth}%
	\includegraphics[scale=.85]{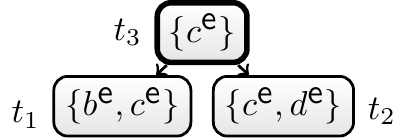}	
	\endminipage
	\vspace{-.95em}
	\caption{Primal graph~$\primal{\prog}$ (left) of~$\prog$, the nested primal graph~$\nested{\prog}{A}$ for~$A=\{b,c,d\}$ (middle) and a tree decomposition~$\mathcal{T}$ for the nested primal graph~$\nested{\prog}{A}$ (right).}
	%
	\label{fig:graph-td2}%
\end{figure}
For our purposes, we require suitable abstractions of~$\primal{\prog}$, given as follows. A \emph{non-epistemic path} in~$\primal{\prog}$ is a path of the form~$a^\ex, v_1^\ax, \ldots, v_l^\ax, b^\ex$ with~$l\geq 0$.  The \emph{nested primal graph~$\nested{\prog}{A}$} over a given set~$A\subseteq\evar(\prog)$ of epistemic atoms is given by~$\nested{\prog}{A}\eqdef (A^\ex, E')$ with~$E'\eqdef \{\{a^\ex,b^\ex\} \mid \{a,b\} \subseteq A, \text{ there is a non-epistemic path from }a^\ex\text{ to }b^\ex\text{ in }\primal{\prog}\}$.

\begin{example}
	Recall program~$\prog$ of Example~\ref{ex:running2}. Figure~\ref{fig:graph-td2} shows the primal graph~$\primal{\prog}$ for program~$\prog$. Given epistemic atoms~$A=\{b,c,d\}$ the nested primal graph~$\nested{\prog}{A}$ can be constructed with edges~$\{b^\ex, c^\ex\}$ and~$\{c^\ex, d^\ex\}$ through any of the non-epistemic paths between the two correlating 
	vertices in~$\primal{\prog}$. 
\end{example}

Indeed, in this section we use the nested primal graph~$\nested{\prog}{A}$ for applying DP in a nested fashion. There, the nested primal graph provides sufficient abstractions of the primal graph, where we count plausible WVIs over~$A$, similar to Listing~\ref{alg:prim}. 
These plausible WVIs over~$A$ are then subsequently extended and refined (to obtain world views), since in each node of a tree decomposition, one chooses again an abstraction~$A'$ that decides on remaining epistemic atoms until all epistemic atoms are considered.
So, if in the beginning we decide that~$A{=}\evar(\prog)$, we end up with full DP and zero nesting, whereas setting~$A{=}\emptyset$ results in full nesting, i.e., no DP. 
%
%
Before we discuss how to choose such a set~$A$ somewhere between these two extreme cases, we define how the ELP that is subject to nesting looks like.
To formalize this, we assume a TD~$\TTT=(T,\chi)$ of~$\nested{\prog}{A}$ and say a set~$U\subseteq \uvar(\prog)$ of atoms is \emph{compatible} with a node~$t$ of~$T$, and vice versa,
if 
\vspace{-.5em}
\begin{enumerate}
	\item[(I)] there is a connected component~$C$ of graph~$\primal{\prog}-A^\ex$ such that~$U=\{a\mid \{a^\ex,a^\ax\}\cap C\neq\emptyset\}$;
	\item[(II)] all neighbor vertices of~$C$ in~$\primal{\prog}$ that are in~$A^\ex$, are contained in~$\chi(t)$, i.e.,
 $\{a^\ex\mid a\in A, u\in U,\text{ there is a }\allowbreak\text{non-epistemic path from }u^\ex\text{ to }a^\ex\text{ in }\primal{\prog}\}\subseteq\chi(t)$.
\end{enumerate}

%

If such a set~$U\subseteq \uvar(\prog)$ of atoms is compatible with a node of~$T$, we say that~$U$ is a \emph{compatible set}.
By construction of the nested primal graph, any atom not in~$A$ is in at least one compatible set, but
a compatible set could be compatible with several nodes of~$T$.
Hence, to enable nested evaluation, we ensure that each nesting atom is evaluated in one unique node~$t$.

As a result, we formalize for every compatible set~$U$ a \emph{unique} node~$t$ of~$T$ that is compatible with~$U$, denoted by $\compat(U)\eqdef t$. 
We denote the union of all compatible sets~$U$ with $\compat(U)=t$, by \emph{nested bag atoms}~$A_t\eqdef \bigcup_{U: \compat(U)=t} U$.
Finally, the \emph{nested bag program}~$\prog_t^A$ for a node $t$ of
$T$, i.e., the ELP subject to nesting, equals $\prog_t^A\eqdef \{r\in \prog \mid \var(r)\subseteq A_t, \evar(r)\subseteq A_t \cup \{a\mid a^\ex\in\chi(t)\}\} \setminus \prog_t$. 
Observe that the definition of nested bag programs ensures that any connected component~$U$ of $\primal{\prog} -A^\ex$ ``appears''
among nested bag atoms of some unique node of~$T$.
Consequently, for each atom $a \in \uvar(\prog) \setminus A$
there is a \emph{unique} node~$t$ such that~$a\in\uvar(\prog_t^A)$.

\begin{example}\label{ex:running8}%
	Considering program~$\prog$ from Example~\ref{ex:running2} and the nested primal graph~$\nested{\prog}{A}$ for~$A=\{b,c,d\}$, Figure~\ref{fig:graph-td2} shows a corresponding TD~$\mathcal{T}$ for the nested primal graph~$\nested{\prog}{A}$. When removing vertices~$A^\ex$ from~$\primal{\prog}$ one can identify the two connected components~$\{a^\ax, b^\ax, a^\ex\}$ and ~$\{c^\ax, d^\ax\}$ each of which building a compatible set in the form of~$U_1\eqdef\{a,b\}$ uniquely compatible with node~$t_1$ and~$U_2\eqdef\{c,d\}$ uniquely compatible with node~$t_2$, i.e., $\compat(U_1)=t_1$ and~$\compat(U_2)=t_2$. Then nested bag programs~$\prog_{t_1}^A = \{r_{1},r_{4},r_{5},r_{8},r_{9},r_{10},r_{11}\}$ and~$\prog_{t_2}^A = \{r_{2},r_{3},r_{6},r_{7},r_{12}\}$ emerge from~$A_{t_1}=\{a,b\}$ and~$A_{t_2}=\{c,d\}$, respectively. Note that~$\prog_{t_3}^A = \emptyset$ because of~$A_{t_3}=\emptyset$.
\end{example}

\longversion{
\begin{algorithm}[t]%
  \KwData{%
     Table algorithm $\algo{H}$, nesting $\depth\geq 0$, epistemic logic program~$\prog$, a set~$A\subseteq \evar(\prog)$ of epistemic atoms, 
and a TD~$\TTT=(T,\chi)$ of the nested primal graph~$\nested{\prog}{A}$ with~$n=\rootOf(T)$. 
  }%
  \KwResult{%
    The number~$\cELP(\prog \sqcup W,\emptyset)$ of world views, obtained by interpreting the table for root~$n$. 
  } %

  $\ATab{\algo{H}} \leftarrow \{\}\qquad \tcc*[h]{empty mapping}$

  \For{\text{\normalfont iterate} $t$ in $\post(T)$}{
    %
    %
    
    $\Tab{} \leftarrow \langle\ATab{\algo{H}}[t_1],\ldots, \ATab{\algo{H}}[t_\ell]\rangle\text{ where }\children(t)=\langle t_1,\ldots,t_\ell \rangle\hspace{-5em}$
  %
  
  $\ATab{\algo{H}}[t]  \leftarrow {\algo{H}}(\depth,\chi(t), \prog_t^A, W, \Tab{})$
%
  }%
  \Return{$\sum_{\langle I,c \rangle\in \ATab{\algo{H}}[n]}c$} 
 \caption{Algorithm ${\adpa}_{\algo{H}}(\depth, \prog, W, A,\TTT)$ for computing the number of world views of~$\prog$ via hybrid dynamic programming on
   TD~${\TTT}$.}
\label{fig:adpontd}
\end{algorithm}%
}

\subsubsection*{Nested Dynamic Programming for ELPs}

Next, we discuss \emph{nested dynamic programming (nested DP)} in order to count world views of an ELP~$\prog$. 
Thereby we aim at solving the more elaborated problem
 $\cELP(\prog \sqcup W, \emptyset)$ for a world view interpretation~$W$ over a set~$X\subseteq\var(\prog)$ of atoms of~$\prog$. This problem amounts to counting the number of world views of~$\prog$ that agree with~$W$ over atoms~$X$.
Hence, we consider a more fine-grained variant of counting world views that for the special case of~$X=\emptyset$ actually coincides with~$\cELP(\prog, \emptyset)$ as stated in Definition~\ref{def:counting}.

Our algorithm for nested dynamic programming, called \hdpa, is presented in Listing~\ref{fig:hdpontd} and relies on the nested primal graph that is utilized in a nested fashion. Therefore, Algorithm~\hdpa takes as first argument an integer for the nesting depth, the ELP~$\prog$ and the WVI~$W$.
Listing~\ref{fig:hdpontd} consists of four separated blocks. 
The first block (Lines~\ref{line:evars}--\ref{line:sat}) comprises solving the base case where~$\prog$ has no epistemic atoms, i.e., no epistemic ``decisions'' are left for solving~$\prog$.
There, if all atoms of~$X$ appear positively or negatively in~$W$, we use two ASP solver calls to check Conditions (1) or (2)+(3) of Defnition~\ref{def:compat}, respecitvely.
Otherwise all four conditions of Definition~\ref{def:compat} are verified via one ELP solver call.
The next block consists of Lines~\ref{line:dec}--Lines~\ref{line:easp}, which computes a tree decomposition~$\mathcal{T}$ of the primal graph of~$\prog$ (nested primal graph with~$A=\evar(\prog)$). Then this block utilizes standard ELP solvers in case $\width(\mathcal{T})$ is out of reach ($\text{threshold}_{\text{hybrid}}$) or nesting is already too deep ($\text{threshold}_{\text{depth}}$).
If this is not the case and $\width(\mathcal{T})$ is insufficient for DP ($\text{threshold}_{\text{abstr}}$), the third block consisting of Lines~\ref{line:nonesting}--\ref{line:decomposenest} chooses a suitable abstraction~$A$ and computes a TD~$\mathcal{T}$ of the nested primal graph~$\nested{\prog}{A}$. 
Finally, the last block comprises of the remaining lines of Listing~\ref{fig:hdpontd}, which performs DP on the TD~$\mathcal{T}$ that is obtained either in Block 2 or Block 3 and returns the solution in Line~\ref{line:nested}. 
The actual recursion (nesting) is via table algorithm~$\algEASP$ that is used during DP in Line~\ref{line:hdp}, discussed next. 

\begin{algorithm}[t]%
  \KwData{%
    Nesting~$\depth \geq 0$, epistemic logic program~${\prog}$, and a WVI~$W$ over a set~$X\subseteq\var(\prog)$ of atoms.\hspace{-2em}} 
  \KwResult{The number~$\cELP(\prog \sqcup W, \emptyset)$ of world views.
  }%
  %
    %

  \smallskip
  



  $A \leftarrow \evar(\prog)$\label{line:evars}


  \If{$A=\emptyset \qquad \tcc*[h]{No Epistemic Decisions left; Verify Decisions}$}{\lIf{$\{a\in X\mid a\notin W, \neg a \notin W\}=\emptyset$}{\Return{$\Card{\answersets{\prog}}=1$ and $\Card{\answersets{\prog \cup \{\leftarrow W\}}}=0$\quad $\tcc*[h]{ASP}\hspace{-5em}$}\label{lab:asp}}\lElse{\Return{$\wvs{\prog \sqcup W} \neq \emptyset$ \qquad$\tcc*[h]{Verify via Standard ELP Solver}$}}}\label{line:sat}
  \vspace{-.15em}
  
  \smallskip
  $\mathcal{T}=(T,\chi) \leftarrow  \text{Decompose}(\primal{\prog})\qquad\tcc*[h]{Decompose via Heuristics}\hspace{-1em}$\label{line:dec}
  
  \If{$\width(\mathcal{T}) \geq\text{threshold}_{\text{hybrid}} \text{ or } \depth \geq \text{threshold}_{\text{depth}}$\qquad\,$\tcc*[h]{\hspace{-.25em}Standard ELP Solver}$}{\label{line:depth}
  \Return{$\cELP(\prog\sqcup W, \emptyset)$}\label{line:easp}\vspace{-.2em}
  }
 
  \smallskip\smallskip
  
  \If{$\width(\mathcal{T}) \geq \text{threshold}_{\text{abstr}}\,\qquad \tcc*[h]{\hspace{-.25em}Abstract \& Decompose via Heuristics \hspace{-.15em}}\,$}{\label{line:nonesting}
  	$A\qquad\quad\,\,\,\,\,\,\,\,\,\,\leftarrow \text{Choose-Abstraction}(A,\prog)$\label{line:abstraction}
  	
	\vspace{-0.2em}$\mathcal{T}=(T,\chi) \leftarrow \text{Decompose}(\nested{\prog}{A})$\label{line:decomposenest}\vspace{-.2em}
  }
  
  \smallskip\smallskip

  
  \For{\text{\normalfont iterate} $t$ in $\post(T)$\qquad\qquad$\tcc*[h]{Dynamic Programming}$}{
    %
    %
    
    $\Tab{} \leftarrow \langle\tab{t_1},\ldots, \tab{t_\ell}\rangle\text{ where }\children(t)=\langle t_1,\ldots,t_\ell \rangle\hspace{-5em}$
  %
  
  $\tab{t}  \leftarrow {\algEASP}(\depth,\chi(t), \prog_t,\prog_t^A, W, \Tab{})$\label{line:hdp}
%
  }%
  \Return{$\sum_{\langle I,c \rangle\in \tab{\rootOf(T)}}c$ \qquad\qquad\qquad $\tcc*[h]{Return Total Count}$} 
\label{line:root}\label{line:nested}
  
%
%
%

  
%
%
\caption{%
    Algorithm $\hdpa(\depth, {\prog}, W)$ 
    for world view counting 
    by means of nested DP. 
  %
}%
\label{fig:hdpontd}%
\end{algorithm}%

The table algorithm~$\algEASP$ is given in Listing~\ref{alg:eprim}. Compared to Listing~\ref{alg:prim}, 
we have two additional parameters, namely the nested bag program and WVI~$W$. 
The main differenc is in Line~\ref{line:introduce} of Listing~\ref{alg:eprim}, where an additional recursive call to~\hdpa is performed.
This recursive call increases the $\depth$ and concerns about the nested bag program that is simplified by the current WVI~$J$ and aims at verifying 
WVI~$W\cup J$ restricted to those atoms that appear also in non-epistemic atoms of a rule of the nested bag program. The other atoms not appearing in such a rule 
will be checked in the context of an other bag.
Intuitively, the resulting count~$c'$ of the recursive call needs to be multiplied as it concerns different epistemic atoms, cf. Line~\ref{line:introduceold} of Listing~\ref{alg:eprim}.

\begin{figure}[t]\vspace{-.5em}%
	\centering %
	\includegraphics{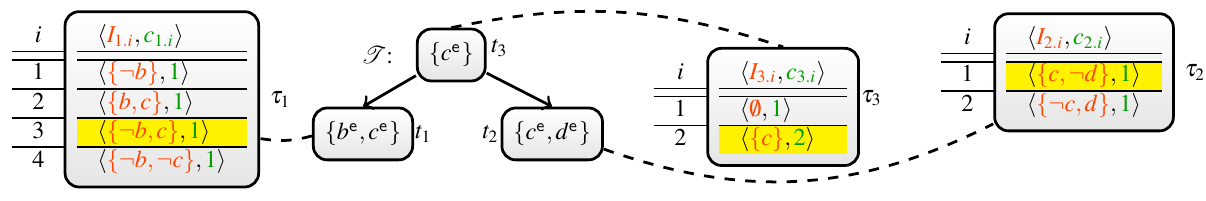}
	\vspace{-1em}
	\caption{
		A TD~$\TTT$ of the nested primal graph~$\nested{\prog}{A}$ of
		program~$\prog$ from Example~\ref{ex:running2} for~$A=\{b,c,d\}$ as well as
		selected tables obtained by~$\#\algo{ELP}$ on~$\Pi$ and~$\TTT$.}
	\label{fig:running8}
\end{figure}%

\begin{example}
Recall program~$\prog$, set~$A$ of epistemic atoms, TD~$\TTT$ of nested primal graph~$\nested{\prog}{A}$ and nested bag programs given in Example~\ref{ex:running8}. Figure~\ref{fig:running8} illustrates computation results obtained during post-order traversal
of~$\TTT$ by~$\#\algo{ELP}$. 
Notice that similar to~$\#\algo{PWV}$ the algorithms enforces the entailment of~$\prog_t$ for each guess, reducing the number of rules for the actual nested call, e.g.~the nested call for node~$t_1$ will only include rules~$\{r_{1},r_{4},r_{5},r_{8},r_{10},r_{11}\}$, c.f.~Example~\ref{ex:running8}. Further observe that while guessing introduced epistemic atoms as in node~$t_1$ and~$t_2$, the epistemic reduct is built over all guessed atoms, but the guess of $c$ is only checked actively in node~$t_2$ using epistemic constraints. Since joining the nodes naturally enforces agreeing assignments of~$c$ this is indirectly checked for~$t_1$.
Similar to Example~\ref{ex:sat}, one can identify that epistemic program~$\prog$ has three world views which can be reconstructing joining agreeing assignments of the tables in-order. For example, we obtain the (incomplete) world view~$\{b,c,\neg d\} = I_{3.2} \cup I_{1.3} \cup I_{2.1}$, as highlighted in yellow.
\end{example}

\begin{algorithm}[t]
  \KwData{Nesting $\depth\geq 0$, bag $\chi_t$, epistemic bag program~$\prog_t$, nested bag program~$\prog_t^A$, world view interpretation~$W$, and sequence
  $\langle \tau_{1},\ldots,\tau_{\ell}\rangle$ of child tables of~$t$.}\KwResult{Table~$\tab{t}.\hspace{-5em}$} 
    \uIf(\hspace{-1em})
  {$\type(t) = \leaf$}{%
\hspace{-1em}$\tab{t} \leftarrow \{ \langle
    \tuplecolor{\inputPredColor}{\emptyset}, \tuplecolor{\epistemiccolor}{1}\rangle\}$\label{row:leaf}
  }%
  \uElseIf{$\type(t) = \intr$ and
    $a^\ex\hspace{-0.1em}\in\hspace{-0.1em}\chi_t$ is introduced}{ %
    \makebox[10.25em][l]{\hspace{-1em}$\tab{t} \leftarrow \{ \langle
      \tuplecolor{\inputPredColor}{J}, \tuplecolor{\epistemiccolor}{c'} \rangle$
    }%
    $|\;\langle \tuplecolor{\inputPredColor}{I},\tuplecolor{\epistemiccolor}{c}\rangle \in \tau_{1},{J\in\{I, I\cup\{a\}, I\cup\{\neg a\}\}}, J \models_{\mathsf{p}} \prog_t,$ \label{line:introduceold}
    
    $\makebox[13.5em]{}\progASP=(\prog_t^A)^J, c'=c\cdot\hdpa(\depth+1, \progASP, (W\cup J)_{|\var(\progASP)}), 
    c'>0\} \hspace{-5em}$\label{line:introduce}\\
    \vspace{-0.05em}
  }\vspace{-0.05em}%
  \uElseIf{$\type(t) = \rem$ and $a^\ex \not\in \chi_t$ is removed}{%
    \makebox[3.227cm][l]{\hspace{-1em}$\tab{t} \leftarrow \{ \langle
      \tuplecolor{\inputPredColor}{I'}, \tuplecolor{\epistemiccolor}{\sum_{\langle J, c' \rangle\in\tau_1: I'\subseteq J} c'}\rangle$}$|\;\langle \tuplecolor{\inputPredColor}{I}, \tuplecolor{\epistemiccolor}{c}\rangle\in\tau_{1}, I'=I\setminus\{a,\neg a\}\}\hspace{-10em}$\hspace{-10em}\vspace{-0.1em}%
} %
  \uElseIf{$\type(t) = \join$}{%
    {
    \hspace{-1em}$\tab{t} \leftarrow \{\langle \tuplecolor{\inputPredColor}{I}, \tuplecolor{\epistemiccolor}{c_1\cdot c_2}\rangle$}$ \; \mid \langle \tuplecolor{\inputPredColor}{I}, \tuplecolor{\epistemiccolor}{c_1}\rangle\in\tau_{1}, \langle \tuplecolor{\inputPredColor}{I}, \tuplecolor{\epistemiccolor}{c_2}\rangle\in\tau_{2}\}\hspace{-5em}$\label{line:join}
    \vspace{-0.15em}
  } %
  \vspace{-0.1em}
  \Return $\tab{t}$ \vspace{-0.25em}
  \caption{Table algorithm~$\algEASP(\depth,\chi_t,\prog_t,\prog_t^A,W,\langle \tau_{1}, \ldots,\tau_\ell\rangle)$ for Counting WVIs.} 
  \label{alg:eprim}
    %
\end{algorithm}


Having established an algorithm for counting, we only briefly discuss how to extend the table algorithm of Listing~\ref{alg:eprim} for \emph{probabilistic world view acceptance} of a WVI (query)~$Q$ via Definition~\ref{def:prob}.
To this end, instead of storing only a WVI and a counter, the rows of the tables of the obtained table algorithm~$\algQELP$ are of the form~$\langle I, c, q\rangle$, where~$I$ is a WVI and~$c$ as well as~$q$ are counters. 
Thereby, $I$ and~$c$ are maintained as before and~$q$ is computed similarly to~$c$, but in Line~\ref{line:introduce} the recursive call for obtaining~$q'$ involves the nested bag program extended by~$Q$, i.e., ${\prog}_t^{A}\sqcup Q$.
Then, instead of summing up counters~$c$ in Line~\ref{line:root} of Listing~\ref{fig:hdpontd}, these adapted tables computed by~$\algQELP$ explained above are used to sum up fractions~$\frac{q}{c}$, which leads the desired result.
Detailed algorithms for $\algQELP$ and~$\hdpa_{\algQELP}$ are depicted in the appendix, cf.\ Listings~\ref{alg:eprim2} and~\ref{fig:hdpontd2}.
%


\section{Implementation \& Preliminary Experiments}

We implemented the algorithm \hdpa, resulting in the solver \nestelp%
\footnote{The solver \nestelp is open source and available at~\href{https://github.com/viktorbesin/nestelp}{github.com/viktorbesin/nestelp}.}%
, which is  written
in Python3. It is based on the system \nesthdb that was presented for variants of model counting~\cite{HecherThierWoltran20}.
For manipulating tables during DP, \nestelp  uses the open source database Postgres~12, which supports instant parallelization and was run on a tmpfs-ramdisk as intended by \nesthdb.
In order to compute TDs (Lines~\ref{line:dec} and~\ref{line:decomposenest} of Listing~\ref{fig:hdpontd}), we use \htd~\cite{AbseherMusliuWoltran17a}, which for every instance outputs TDs of decent widths in a runtime below some seconds.
For solving decision problems of logic programs in Line~\ref{lab:asp} we used \clingo~5.4\longversion{~\cite{GebserEtAl19}}.
For solving ELP problems in Lines~\ref{line:sat} and~\ref{line:easp}, we utilized \eclingo~0.2.
Internally, we set~$\text{threshold}_{\text{hybrid}}$ $=45$, $\text{threshold}_{\text{abstract}}=8$ and allowed nesting once, \FIX{which overall seemed to produce good results. However, these parameters are not the result of extensive performance tuning, but were chosen as initial values with the goal of balancing abstractions and hybrid (standard) solving.}
\FIX{For finding good abstractions in Line~\ref{line:abstraction}, i.e., searching for epistemic atoms
when constructing the nested primal graph, we employ a logic program similar to \nesthdb.
Intuitively,
we thereby aim for a preferably large set~$A$ of epistemic atoms such that the resulting graph~$N_\Pi^A$ is reasonably sparse.
This is achieved heuristically by minimizing the number of edges of~$N_\Pi^A$.
To this end, we use built-in optimization of \clingo, where we take the best results after running at most 35 seconds.
For the concrete encodings, we refer to the online repository of \nestelp as given above.}
Our implementation supports both world view \emph{counting} as given in Definition~\ref{def:counting} as well as \emph{probabilistic} world view acceptance of Definition~\ref{def:prob}.

\subsection*{Benchmark Setting}

In order to draw conclusions about the efficiency of our implementation, we conducted a series of benchmarks.  All our used benchmark instances, raw results and detailed data are available online at~\href{https://tinyurl.com/iclp21-nestelp}{tinyurl.com/iclp21-nestelp}. 
In our benchmarks we compare wall clock runtime of \nestelp and \eclingo~\cite{CabalarEtAl20}, where a timeout is considered to occur after 1200 seconds and each solver was granted 16GB of main memory (RAM) per run. We restricted our solver to 12 physical cores.
\FIX{In \emph{single core  mode (sc)} of \nestelp, only one physical core was used, which allows us to compare the performance with other single-core solvers}.
%
Benchmarks were conducted on a cluster consisting of 12 nodes. 
Each node of the cluster is equipped
with two Intel Xeon E5-2650 CPUs and each of these 12 physical cores runs
at 2.2 GHz clock speed that has access to 256 GB shared RAM. %
Results are gathered on Ubuntu~16.04.1 LTS OS that is powered on kernel~4.4.0-139. We disabled hyperthreading and used Python 3.7.6.
%
\subsection*{Benchmark Instances}
The following instances are considered from the literature and extended accordingly.

\vspace{-1em}
\paragraph{Classic-Scholarship.} As in previous works~\cite{CabalarEtAl20}, this is a set of 25 non-ground ELP programs encoding the Scholarship Eligibility problem~\cite{aaai:Gelfond91} for one to twenty-five students, where all entities are independent from each other. If a students eligibility is not determined by the plain logic rules, an epistemic rule implies the interview of the student.

\vspace{-1em}
\paragraph{Yale-Shooting.} This is a set of 12 non-ground ELP programs~\cite{CabalarEtAl20} encoding the Yale Shooting problem~\cite{HanksMcdermott1986}. With each instance the knowledge of the initial state, i.e., if the gun is initially loaded or not, is incomplete.

\vspace{-1em}
\paragraph{Large-Scholarship (L-S).} While classic-scholarship is limited to 25 instances, large-scholarship can be configured to a number of students, i.e., a student-wise extension to classic-scholarship. As part of our testing, we implemented a generator for such instances, using existing instances to initialize more students. This set consists of 500 instances ranging from 5 to 2500 students.

\vspace{-1em}
\paragraph{Many-Scholarship (M-S).} In comparison to classic-scholarship, where all students are part of one unique world view, many-scholarship extends the situation and aims for a more relaxed situation, where additionally a students eligibility is ranked with low or high chances. This often results in many world views per student. Our generator is implemented in a way such that both introduced instance sets are supported.
Also this set consists of 500 instances.

\noindent

\subsection*{Benchmark Scenarios}
\noindent
We considered the following three scenarios in order to test the efficiency of~\nestelp.

\begin{itemize}
	\item [S1] Counting world views for the classical-scholarship as well as yale-shooting instances.
	\item [S2] Counting world views for large-scale instances, thereby using large-scholarship and many-scholarship instances. For a fair comparison, we allow \eclingo to decide WV existence.
	\item [S3] Probabilistic reasoning \emph{[pr]} for large-scale instances. This scenario concerns probabilistic WV acceptance using also large-scholarship and many-scholarship instances. 
\end{itemize}

\noindent Based on these scenarios, we state corresponding hypothesis that shall be verified in this section.

\begin{itemize}
	\item [H1] \nestelp is competitive for counting, although monolithic solvers like \eclingo are faster. 
	\item [H2] Our implementation \nestelp is rather competitive for large-scale instances.
	\item [H3] Probabilistic reasoning comes almost for the same cost as counting in the solver \nestelp.
\end{itemize}

\subsection*{Experimental Results}

\begin{figure}
\begin{minipage}{.45\linewidth}
\resizebox{.95\columnwidth}{!}{%
  \begin{tabular}{l|l|rrrHHr|r}
    \toprule
    \multirow{2}{*}{solver} & \multirow{2}{*}{max($\width$)} & \multicolumn{4}{c}{\#solved ($\width$ range)} & && \multirow{2}{*}{time[h]}\\
&& 0-5 & 5-20 & $>$20 & best & unique & $\sum$ \\
    \midrule
    \multicolumn{9}{c}{Classic-Scholarship: 25 instances}\\
    eclingo & 1.0 & 25 & 0 & 0 & 24 & 0 & 25 & 0.01 \\
    nestelp & 1.0 & 25 & 0 & 0 & 1 & 0 & 25 & 0.01 \\
    nestelp (sc) & 1.0 & 25 & 0 & 0 & 0 & 0 & 25 & 0.02 \\
    \multicolumn{9}{c}{Yale-Shooting: 12 instances}\\
    eclingo & 61.0 & 2 & 3 & 3 & 8 & 0 & 8 & 1.34 \\
    nestelp & 61.0 & 2 & 3 & 3 & 0 & 0 & 8 & 1.37 \\
    nestelp (sc) & 61.0 & 2 & 3 & 3 & 0 & 0 & 8 & 1.40 \\
     \multicolumn{9}{c}{$\Sigma$: 37 instances}\\
    eclingo & 61.0 & 27 & 3 & 3 & 32 & 0 & 33 & 1.35 \\
    nestelp & 61.0 & 27 & 3 & 3 & 1/1 & 0/0 & 33 & 1.39  \\
    nestelp (sc) & 61.0 & 27 & 3 & 3 & 0/0 & 0/0 & 33 & 1.42  \\
    \bottomrule
  \end{tabular}
  }%
\end{minipage}%
\begin{minipage}{.50\linewidth}
\includegraphics[scale=.4]{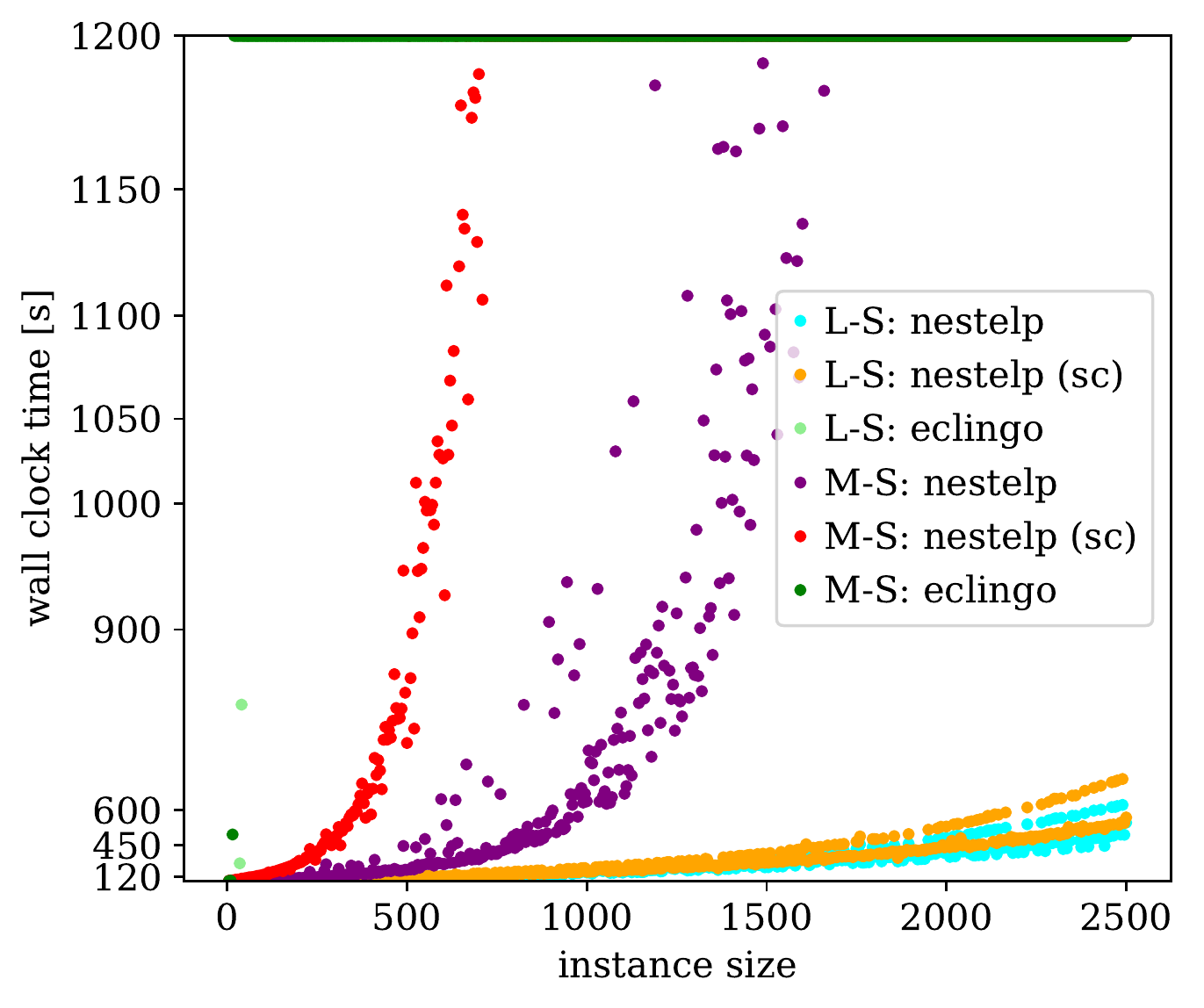}
\end{minipage}\vspace{-.6em}
	\caption{Detailed results (left) over Scenario S1 showing maximal width of the primal graph among solved instances, solved instances over certain width ranges, as well as total runtime in hours, where timeouts count as 1200s. Line plot (right) of instances L-S and M-S for Scenario S2, where instances are ordered ascendingly according to instance size.}
	\label{lab:s1s2}
\end{figure}

\begin{figure}\vspace{-.75em}
\begin{minipage}{.5\linewidth}
	\includegraphics[scale=.42]{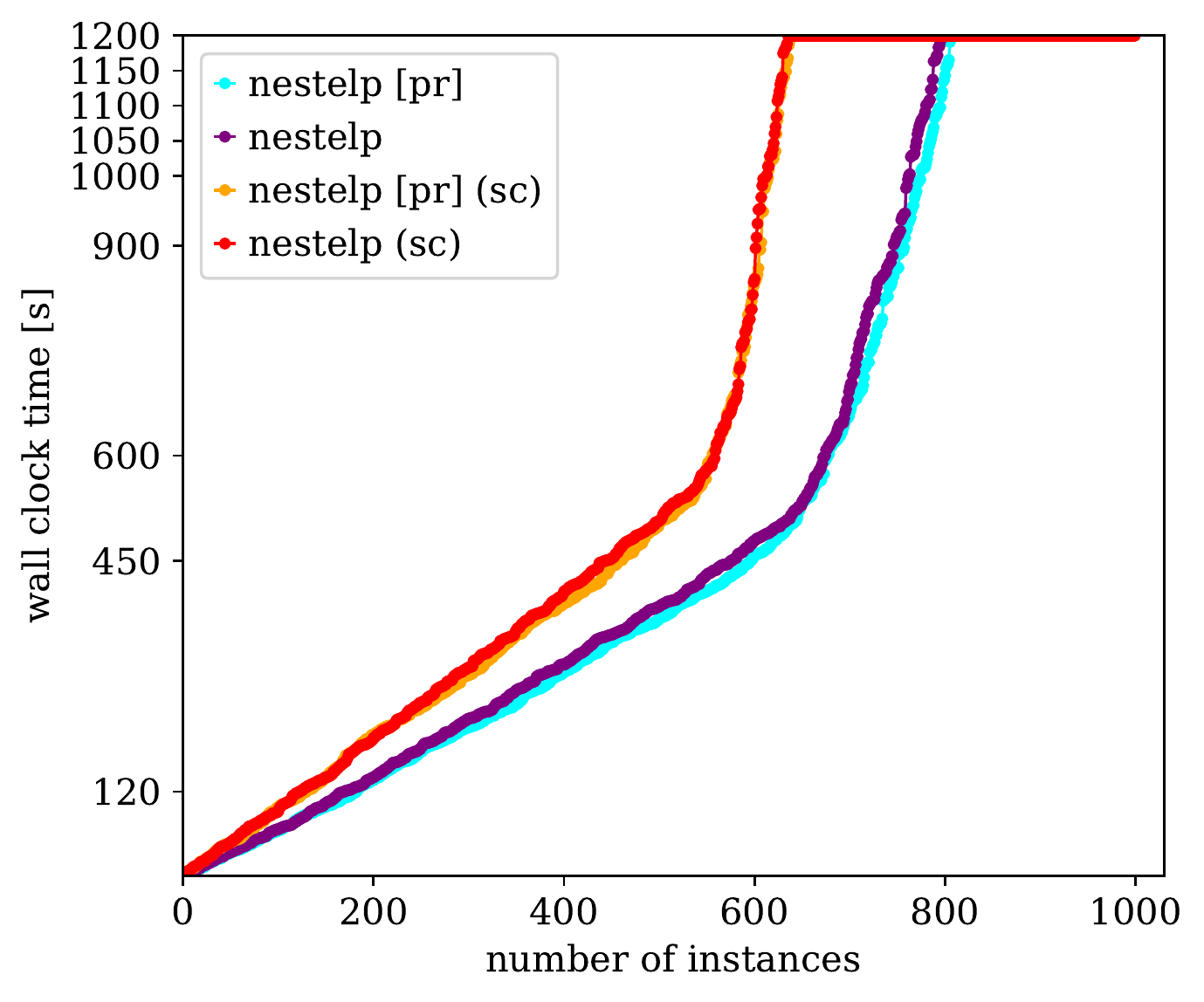}
\end{minipage}
\begin{minipage}{.44\linewidth}
\resizebox{.95\columnwidth}{!}{%
  \begin{tabular}{l|l|HHHllrr}
    \toprule
    solver & max\_width & 0-5 & 5-20 & $>$20 & \#fastest & \#unique & \#solved & time[h] \\
    \midrule
    \multicolumn{9}{c}{Large-Scholarship (L-S): 500 instances}\\
    nestelp [pr] & 1.0 & 500 & 0 & 0 & 275 & 0 & 500 & 35.72 \\
    nestelp & 1.0 & 500 & 0 & 0 & 225 & 0 & 500 & 36.50 \\
    nestelp [pr] (sc) & 1.0 & 500 & 0 & 0 & 4 & 0 & 500 & 39.46 \\
    nestelp (sc) & 1.0 & 500 & 0 & 0 & 4 & 0 & 500 & 40.08 \\
    eclingo & 1.0 & 8 & 0 & 0 & 5 & 0 & 8 & 164.32 \\
    \multicolumn{9}{c}{Many-Scholarship (M-S): 500 instances}\\
    nestelp [pr] & 2.0 & 306 & 0 & 0 & 183 & 18 & 306 & 106.14 \\
    nestelp & 2.0 & 296 & 0 & 0 & 132 & 9 & 296 & 109.19 \\
    nestelp [pr] (sc) & 2.0 & 138 & 0 & 0 & 0 & 0 & 138 & 142.61 \\
    nestelp (sc) & 2.0 & 135 & 0 & 0 & 0 & 0 & 135 & 143.31 \\
    eclingo & 2.0 & 3 & 0 & 0 & 1 & 0 & 3 & 165.81 \\
        \multicolumn{9}{c}{$\Sigma$: 1000 instances}\\
    nestelp [pr] & 2.0 & 806 & 0 & 0 & 458 & 18 & 806 & 141.87  \\
    nestelp & 2.0 & 796 & 0 & 0 & 357 & 9 & 796 & 145.70  \\
    nestelp [pr] (sc) & 2.0 & 638 & 0 & 0 & 4 & 0 & 638 & 182.07 \\
    nestelp (sc) & 2.0 & 635 & 0 & 0 & 4 & 0 & 635 & 183.39  \\
    eclingo & 2.0 & 11 & 0 & 0 & 6 & 0 & 11 & 330.13 \\
    \bottomrule
  \end{tabular}
  }%
\end{minipage}
	\caption{Scenario S3: Cactus plot (left), whose x-axis shows the number of instances; the y-axis depicts runtime sorted ascendingly for each solver individually. Detailed results (right).}
	\label{lab:s3}
\end{figure}
The results for Scenario S1 in comparison with \eclingo are summarized in the table of Figure~\ref{lab:s1s2}. Overall it can be seen that \nestelp can keep up with a traditional solver like \eclingo, but, as expected, \nestelp introduces additional overhead by the creation of tables and the general build-up for dynamic programming. Small instances, as for S1, do not benefit from that process, that is why we expected such results. The number of solved instances is the same for both systems, overall agreeing with our Hypothesis H1.
The line plot in Figure~\ref{lab:s1s2} shows an outstanding performance of \nestelp for instances L-S and even M-S. Both instance sets allow their instances to be arranged into decompositions with low treewidth, representing instances where \nestelp can exploit all its features. Further it can be seen that parallelism of \nestelp has better performance than the single-core experiments
(\nestelp (sc)), 
indicating that there are enough independent nodes such that parallelism is beneficial. Even with the fair comparison to \eclingo, the solver \nestelp proves its ability to handle large-scale instances well, as proposed in Hypothesis H2. 
As it can be seen in the cactus plot in Figure~\ref{lab:s3}, the effort needed for probabilistic reasoning is very small in comparison to world view counting. Since \nestelp intuitively only processes sub-calls where they are justified, i.e.,~only when there are any world views, there is little to no difference in the plot. While agreeing with Hypothesis H3, we even believe that the visible differences are due to scattering factors like query optimization and CPU clocking.
To summarize, the systems performance can be described quite competitively with a higher number of solved instances in similar or even shorter runtimes. Furthermore, consider that \nestelp uses \eclingo for sub-calls, leading to the assumption that every revision of the base solver will improve our system too.
\section{Conclusion}
In this work we studied counting world views of epistemic logic programs (ELPs) and extended this further to probabilistic reasoning. We took up ideas of a theoretical algorithm that utilizes treewidth and progressively turned this into an efficient solver. Our solver 
\nestelp 
works on iteratively computing and refining (graph) abstractions of the ELP and counting world views over epistemic atoms of the abstract program. Then, the count is subsequently improved by refining the abstraction in a nested fashion, for which we use our algorithm or existing (E)LP solvers.
Specifically for counting and probabilistic reasoning, 
\nestelp  seems to scale well.
For future work we plan on further optimizing this technique, which however automatically improves with the availability of faster solvers as those are the core engines in \nestelp. 
Further, given recent insights on complexity results for treewidth, e.g.,~\cite{FichteHecherPfandler20,FichteHecherMeier21}, the techniques developed and applied in this work could be also carried out for other formalisms like abstract argumentation or description logics.

\section*{Acknowledgements}

This work has been supported by the Austrian Science Fund (FWF),
   Grants P32830 and Y698, as well as the Vienna Science and Technology Fund, Grant WWTF ICT19-065. We would like to thank the reviewers for their detailed and valuable comments. Part of the research was carried out while Hecher was visiting the Simons Institute for the Theory of Computing.

\bibliographystyle{acmtrans}
{

\bibliography{references}
}
\appendix

\clearpage
\section{Probabilistic Reasoning}

\begin{algorithm}[h]
  \KwData{Nesting $\depth\geq 0$, bag $\chi_t$, epistemic bag program~$\prog_t$, nested bag program~$\prog_t^A$, world view interpretation~$W$, WVI (query)~$Q$, and sequence
  $\langle \tau_{1},\ldots,\tau_{\ell}\rangle$ of child tables of~$t$.{~\bf Out:} Table~$\tab{t}.\hspace{-5em}$} 
    \lIf(\hspace{-1em})
  {$\type(t) = \leaf$}{%
    $\tab{t} \leftarrow \{ \langle
    \tuplecolor{\inputPredColor}{\emptyset}, \tuplecolor{\epistemiccolor}{1}, \tuplecolor{\specialPredColor}{1} \rangle\}$\label{row2:leaf}
  }%
  \uElseIf{$\type(t) = \intr$ and
    $a^\ex\hspace{-0.1em}\in\hspace{-0.1em}\chi_t$ is introduced}{ %
    \makebox[10.25em][l]{\hspace{-1em}$\tab{t} \leftarrow \{ \langle
      \tuplecolor{\inputPredColor}{J}, \tuplecolor{\epistemiccolor}{c'}, \tuplecolor{\specialPredColor}{q'} \rangle$
    }%
    $|\;\langle \tuplecolor{\inputPredColor}{I},\tuplecolor{\epistemiccolor}{c}, \tuplecolor{\specialPredColor}{q}\rangle \in \tau_{1},{J\in\{I, I\cup\{a\}, I\cup\{\neg a\}\}}, J \models_{\mathsf{p}} \prog_t,$ \label{line2:introduceold}
    
    $\makebox[11.5em]{}\progASP=(\prog_t^A)^J, c'=c\cdot\hdpa(\depth+1, \progASP, (W\cup J)_{|\var(\progASP)}), 
    c'>0\} \hspace{-5em}$\label{line2:introduce}\\
    
     $\makebox[11.5em]{}\progASP'=(\prog_t^A\sqcup Q)^J, q'=q\cdot\hdpa(\depth+1, \progASP', (W\cup J)_{|\var(\progASP')}), 
    q'>0\} \hspace{-5em}$\label{line2:introduce2}
    \vspace{-0.05em}
  }\vspace{-0.05em}%
  \uElseIf{$\type(t) = \rem$ and $a^\ex \not\in \chi_t$ is removed}{%
    \makebox[6cm][l]{\hspace{-1em}$\tab{t} \leftarrow \{ \langle
      \tuplecolor{\inputPredColor}{I'}, \tuplecolor{\epistemiccolor}{\sum_{\langle J, c', q' \rangle\in\tau_1: I'\subseteq J} c'}, \tuplecolor{\specialPredColor}{\sum_{\langle J, c', q' \rangle\in\tau_1: I'\subseteq J} q'}\rangle $}$|\;\langle \tuplecolor{\inputPredColor}{I}, \tuplecolor{\epistemiccolor}{c}, \tuplecolor{\specialPredColor}{q}\rangle\in\tau_{1}, I'=I\setminus\{a,\neg a\}\}\hspace{-10em}$\hspace{-10em}\vspace{-0.1em}%
} %
  \ElseIf{$\type(t) = \join$}{%
    \makebox[10.25em][l]
    {
    \hspace{-1em}$\tab{t} \leftarrow \{\langle \tuplecolor{\inputPredColor}{I}, \tuplecolor{\epistemiccolor}{c_1\cdot c_2}, \tuplecolor{\specialPredColor}{q_1\cdot q_2}\rangle$}$ \mid \langle \tuplecolor{\inputPredColor}{I}, \tuplecolor{\epistemiccolor}{c_1}, \tuplecolor{\specialPredColor}{q_1}\rangle\in\tau_{1}, \langle \tuplecolor{\inputPredColor}{I}, \tuplecolor{\epistemiccolor}{c_2}, \tuplecolor{\specialPredColor}{q_2}\rangle\in\tau_{2}\}\hspace{-5em}$\label{line2:join}
    \vspace{-0.15em}
  } %
  \vspace{-0.4em}
  \caption{Table algorithm~$\algQELP(\depth,\chi_t,\prog_t,\prog_t^A,W,Q,\langle \tau_{1}, \ldots,\tau_\ell\rangle)$ for nice TDs of the nested primal graph representation.}
  \label{alg:eprim2}
    %
\end{algorithm}

\begin{algorithm}[h]%
  \KwData{%
    Nesting~$\depth \geq 0$, epistemic logic program~${\prog}$, WVI~$W$  over a set~$X\subseteq\var(\prog)$ of atoms, and WVI (query)~$Q$.\hspace{-2em}} 
  \KwResult{The probability~$\prob(\prog \sqcup W, Q)$ of~$Q$ being compatible with a world view.
  }%
  %
    %

  \smallskip
  



  $A \leftarrow \evar(\prog)$\label{line2:evars}


  \If{$A=\emptyset \qquad \tcc*[h]{No Epistemic Decisions left; Verify Decisions}$}{\lIf{$\{a\in X\mid a\notin W, \neg a \notin W\}=\emptyset$}{\Return{$\Card{\answersets{\prog}}=1$ and $\Card{\answersets{\prog \cup \{\leftarrow W\}}}=0$\quad $\tcc*[h]{ASP}\hspace{-5em}$}\label{lab2:asp}}\lElse{\Return{$\wvs{\prog \sqcup W} \neq \emptyset$ \qquad$\tcc*[h]{Verify via Standard ELP Solver}$}}}\label{line2:sat}
  \vspace{-.15em}
  
  \smallskip
  $\mathcal{T}=(T,\chi) \leftarrow  \text{Decompose}(\primal{\prog})\qquad\tcc*[h]{Decompose via Heuristics}\hspace{-1em}$\label{line2:dec}
  
  \If{$\width(\mathcal{T}) \geq\text{threshold}_{\text{hybrid}} \text{ or } \depth \geq \text{threshold}_{\text{depth}}$\qquad\,$\tcc*[h]{\hspace{-.25em}Standard ELP Solver}$}{\label{line2:depth}
  \Return{$\prob(\prog\sqcup W, Q)$}\label{line2:easp}\vspace{-.2em}
  }
 
  \smallskip\smallskip
  
  \If{$\width(\mathcal{T}) \geq \text{threshold}_{\text{abstr}}\,\qquad \tcc*[h]{\hspace{-.25em}Abstract \& Decompose via Heuristics \hspace{-.15em}}\,$}{\label{line2:nonesting}
  	$A\qquad\quad\,\,\,\,\,\,\,\,\,\,\leftarrow \text{Choose-Abstraction}(A,\prog)$\label{line2:abstraction}
  	
	\vspace{-0.2em}$\mathcal{T}=(T,\chi) \leftarrow \text{Decompose}(\nested{\prog}{A})$\label{line2:decomposenest}\vspace{-.2em}
  }
  
  \smallskip\smallskip

  
  \For{\text{\normalfont iterate} $t$ in $\post(T)$\qquad\qquad$\tcc*[h]{Dynamic Programming}$}{
    %
    %
    
    $\Tab{} \leftarrow \langle\tab{t_1},\ldots, \tab{t_\ell}\rangle\text{ where }\children(t)=\langle t_1,\ldots,t_\ell \rangle\hspace{-5em}$
  %
  
  $\tab{t}  \leftarrow {\algQELP}(\depth,\chi(t), \prog_t,\prog_t^A, W, Q, \Tab{})$\label{line2:hdp}
%
  }%
  \Return{$\sum_{\langle I,c,q \rangle\in \tab{\rootOf(T)}, c> 0} \frac{q}{c}$ \qquad\qquad\qquad $\tcc*[h]{Return Total Probability}$} 
\label{line2:root}\label{line2:nested}
  
%
%
%

  
%
%
\caption{%
    Algorithm $\hdpa_\algQELP(\depth, {\prog}, W, Q)$ 
    for probabilistic world view acceptance 
    via nested DP. 
  %
}%
\label{fig:hdpontd2}%
\end{algorithm}%

\label{lastpage}
\end{document}